\documentclass[conference]{IEEEtran}
\usepackage{subcaption}
\usepackage{float}
\usepackage{multirow, multicol}
\usepackage{hhline}
\usepackage{amsmath}
\usepackage{array, makecell}%
\usepackage{bm}
\usepackage{amsbsy}
\usepackage{booktabs}
\usepackage{xspace}
\usepackage{xcolor}
\usepackage{newfloat}
\usepackage[export]{adjustbox}
\usepackage{enumitem}
\usepackage[hidelinks]{hyperref}
\usepackage{url}

\usepackage[nottoc]{tocbibind}
\usepackage{mathtools}

\usepackage[utf8]{inputenc} 
\usepackage{csquotes}
\usepackage{enumitem}
\usepackage{ifthen}
\usepackage{lipsum}
\usepackage{graphicx}
\usepackage{caption}
\usepackage{subcaption}
\IEEEoverridecommandlockouts
\def\BibTeX{{\rm B\kern-.05em{\sc i\kern-.025em b}\kern-.08em
    T\kern-.1667em\lower.7ex\hbox{E}\kern-.125emX}}
    
\def\eg{\textit{e.g.,}\@\xspace} 
\def\ie{\textit{i.e.,}\@\xspace} 
 
\def\cf{\textit{cf.}\@\xspace} 
 
\def\wrt{w.r.t.\@\xspace}

\def\faceqnet{FaceQNet\@\xspace}

\begin{document}

\title{Fun Selfie Filters in Face Recognition: \\Impact Assessment and Removal\\
{\large Cristian Botezatu, Mathias Ibsen, Christian Rathgeb, Christoph Busch }
\thanks{The authors are with the Norwegian Biometrics Laboratory at NTNU and the Biometrics and Internet Security Research Group at Hochschule Darmstadt\\
E-mail: cristian.botezatu@ntnu.no}
}
\author{}
\maketitle
\thispagestyle{plain}
\pagestyle{plain}

\begin{abstract}
This work investigates the impact of fun selfie filters, which are frequently used to modify selfies, on face recognition systems. Based on a qualitative assessment and classification of freely available mobile applications, ten relevant fun selfie filters are selected to create a database. To this end, the selected filters are automatically applied to face images of public face image databases. Different state-of-the-art methods are used to evaluate the influence of fun selfie filters on the performance of face detection using dlib, RetinaFace, and a COTS method,  sample quality estimated by \faceqnet  and MagFace, and recognition accuracy employing ArcFace and a COTS algorithm. The obtained results indicate that selfie filters negatively affect face recognition modules, especially if fun selfie filters cover a large region of the face, where the mouth, nose, and eyes are covered. To mitigate such unwanted effects, a GAN-based selfie filter removal algorithm is proposed which consists of a segmentation module, a perceptual network, and a generation module. In a cross-database experiment the application of the presented selfie filter removal technique has shown to significantly improve the biometric performance of the underlying face recognition systems.
\end{abstract}

\vspace{10pt}
\begin{IEEEkeywords}
Biometrics, face recognition, selfie filter, face occlusion,  generative adversarial network, inpainting
\end{IEEEkeywords}

\begin{figure}[!t]
\centering
\includegraphics[width=0.90\linewidth]{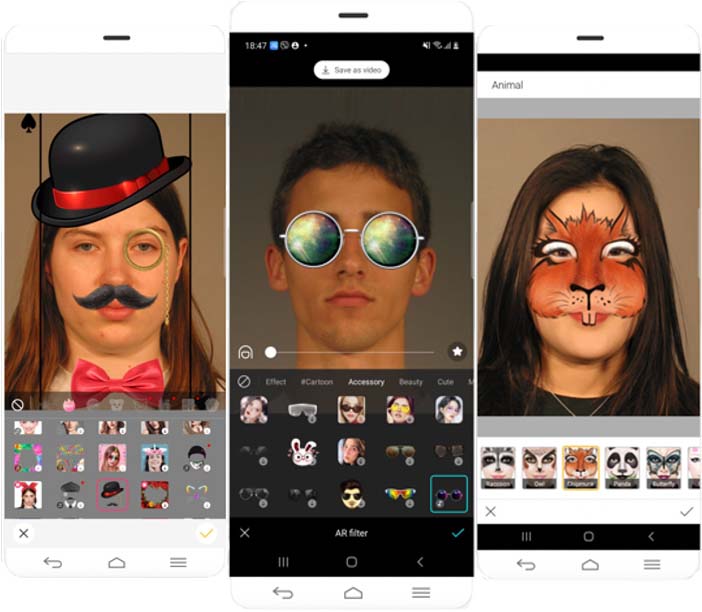}
\caption{Examples of fun selfie filters applied to face images. }\label{fig:funfilters}\vspace{-0.3cm}
\end{figure}

\ifCLASSOPTIONcompsoc
\IEEEraisesectionheading{\section{Introduction}\label{sec:introduction}}
\else
\section{Introduction}
\label{sec:introduction}
\fi
In the recent past, the use of deep convolutional neural networks has achieved remarkable improvements in face recognition (FR) accuracy, surpassing human-level performance \cite{Taigman14,Guo-DeepFaceSurvey-2019}. Due to these breakthrough advances FR technologies have become an essential tool for identity management systems and forensic investigations worldwide. In the latter application scenario, public content plays an important role, especially facial images from social media \cite{CNN-SM-Investigations,DetroitPoliceDepartment-WeeklyReportOnFaceRecognition-2021,MichiganLive-HowPoliceMonitorSocialMediaToFindCrimeSuspects-2021}.  However, before sharing their face images on social media platforms, \eg Facebook or Instagram, users frequently edit them to achieve a desired impact. Common editing tools include beautification filters which may apply significant alterations to the facial shape and texture, \eg by enlarging the eyes or smoothing of the skin. Furthermore, so-called \emph{fun selfie filters} are frequently applied by users to add to the amusement, as illustrated in Fig.~\ref{fig:funfilters}, inducing severe alterations and occlusions to face images. 

In a FR system, fun selfie filters applied to face images  
are expected to represent a challenge for various processing stages \cite{Rathgeb-HandbookFaceManipulationDetection-2022, Ibsen-DigitalFaceManipulationsBiometricSystems-Handbook-2022}. For instance, a large coverage of the facial region by a fun selfie filter may hamper face detection or face sample quality estimation. In addition,  biometric comparison scenarios where one of the face images to be compared has been altered with a fun selfie filter are expected to be challenging. However, to the best of the authors' knowledge, the impact of fun selfie filters on the performance of state-of-the-art FR systems has not been investigated yet. 

Recently, deep learning techniques have been applied for the purpose of image inpainting. In particular methods based on generative adversarial network (GAN) have shown impressive results for removing facial occlusion \cite{Zeng21}, \eg caused by medical masks \cite{din2020gan}. In order to perform well, such techniques usually require a large amount of realistic training data  containing face image with and without targeted occlusions. To the best of the authors' knowledge, so far the feasibility of GAN-based removal of fun selfie filters has not been investigated in the scientific literature.

In order to assess the impact of fun selfie filters on FR systems and mitigate their potentially negative effects, this work makes the following contributions:

\begin{itemize}
    \item A qualitative assessment of fun selfie filters, available in mobile application stores, is conducted. Based on this assessment, ten highly relevant filters are identified and classified \wrt the face image alterations.
    \item The automated creation of a dataset generated from images of more than 1,000 subjects of the public FRGCv2 and FERET face image databases.
    \item A comprehensive evaluation of the impact of fun selfie filters on the performance of face detection, face sample quality, and FR, along with a detailed discussion of obtained results.
    \item A selfie filter removal algorithm which is created by adapting existing network architectures for segmentation and inpainting \cite{din2020gan,yu2019gate}, along with a detailed evaluation of FR performance before and after the selfie filter removal.
\end{itemize}

The remainder of this paper is organised as follows: related works are briefly summarised in Sect.~\ref{sec:realted}. The creation of the selfie filter face image dataset is described in detail in Sect.~\ref{sec:db}. Subsequently, the impact of selfie filters on different FR sub-systems is evaluated and discussed in Sect.~\ref{sec:impact}. Sect.~\ref{sec:gan} introduces the novel GAN-based selfie removal which is applied to the created dataset to evaluate to which extent it mitigates the effects of selfie filters. Sect.~\ref{sec:summary} concludes the work.

\section{Related Work}\label{sec:realted}



\label{sec:related_work}
Facial occlusions challenge FR systems which are able to cope with the occlusion problem in three main ways \cite{Zeng21}: 
\begin{itemize}
    \item \textbf{Occlusion Robustness:} apply a patch-based or learning-based feature extraction strategy to describe the feature space that is less affected by facial occlusions (\eg \cite{LBP1} \cite{LBP2} \cite{LBP3} \cite{LBP4} \cite{LBP5}).
    \item \textbf{Occlusion Awareness:} feature extraction methods which detect occulded facial regions and subsequently only employ visible face parts for FR (\eg \cite{LBP6} \cite{LBP7} \cite{LBP8} \cite{LBP9}).
    \item \textbf{Occlusion Recovery:} techniques which aim at reconstructing the occluded face parts prior to applying the feature extraction of the FR systems (\eg \cite{REC1} \cite{REC2} \cite{REC3} \cite{REC4} \cite{eyeglassRemoval}).
\end{itemize}

Many works have reported negative impacts of facial occlusions on FR, \eg caused by sunglasses or face masks \cite{Zeng21}. The common facial occlusions that challenge the current state-of-the-art FR systems are listed in Tab. \ref{occ_tab}.

\begin{table}[h!]
\caption{Categorization of occlusion types \cite{Zeng21}.}
\label{occ_tab}
\centering
\begin{adjustbox}{max width=\columnwidth}
\begin{tabular}{ll}
\toprule
\textbf{Occlusion Scenario} & \textbf{Examples}                                                                                                       \\ \midrule
Facial accessories          & eyeglasses, sunglasses, scarves, mask, hat, hair                                                                        \\ \midrule
External occlusions         & occluded by hands and random objects                                                                                    \\ \midrule
Limited field of view       & partial faces                                                                                                           \\ \midrule
Self-occlusions             & non-frontal pose                                                                                                        \\ \midrule
Extreme illumination        & part of face highlighted, other parts in shadow                 \\ \midrule
Artificial occlusions       & random black or white rectangles.       \\ \bottomrule
\end{tabular}
\end{adjustbox}
\end{table}

Similar to occlusions, strong makeup \cite{Dantcheva12a,Chen14a,Rathgeb-ImpactDetectionFacialBeautificationSurvey-ACCESS-2019} or even facial tattoos and painting \cite{Ibsen-ImpactFacialTattoosPaintingsFaceRecognitionSystems-BMT-2021} have been shown to negatively influence FR systems, especially in cases where significant parts of a face are covered with tattoos or paint.

Additionally, due to the recent COVID-19 pandemic, the trend of wearing facial masks in public is growing all over the world. Some people wear masks to guard themselves from certain viruses, pollution, or to simply hide their face and emotions from the public. However, in some cases facial masks are used intentionally to trick the FR systems. Hence, many research activities focus on algorithms to increase FR performance when dealing with masks that cover a large area of an individual's face \cite{9607384,9484337}.

Focusing on selfie filters, the impact of beautification filters has firstly been analysed in \cite{Ferrara2016}. It was shown that the performance of a FR system might significantly drop in case a beautification filter drastically alters the facial appearance. In more recent works it has been shown that FR systems can be robust to moderate alterations resulting from the use of beautification applications, \eg in \cite{Bharati16,Rathgeb-PRNU-Retouching-Detection-BMT-2020,Rathgeb-DifferentialDetectionRetouching-ACCESS-2020}. With respect to fun selfie filters, the Specs on Faces (SoF) dataset was introduced in \cite{AFIFI201977} for the purpose of evaluating various tasks, \eg face detection and gender prediction, in challenging environmental scenarios. This face database contains face images of 112 subjects to which two fun selfie filters have been applied. However, its small size and the fact that facial images of the said database were mostly captured in a single session makes the SoF dataset less suitable for FR performance evaluations. The large amount of facial occlusion variations, as well as their possible random placement on the face, makes FR under occlusion a yet unresolved issue.

\begin{figure*}[h!]
\centering
\includegraphics[width=1\linewidth]{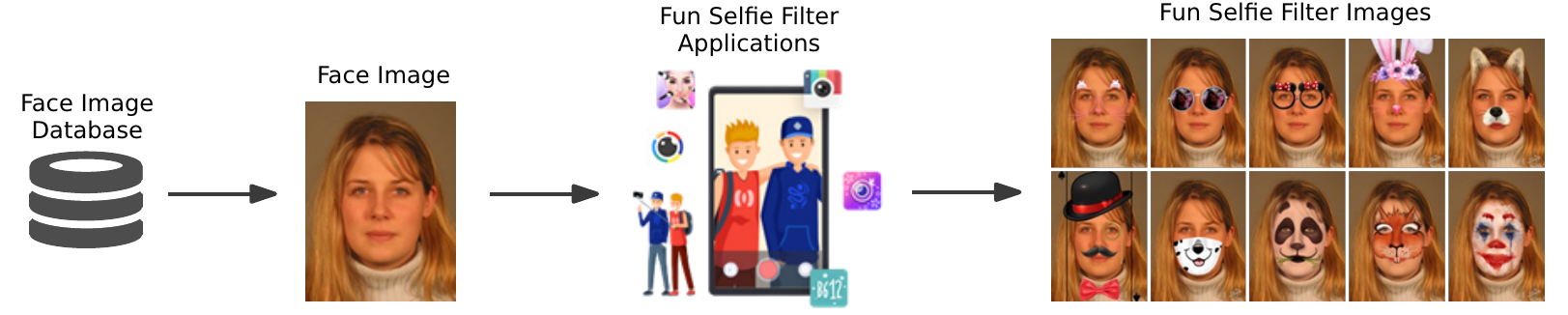}
\caption{Fun selfie filter dataset creation workflow.}
\label{fig:selfie_filter_creation}
\vspace{-10pt}
\end{figure*}

So far, fun selfie filters have not been considered as potential occlusions a FR system has to deal with. Hence, the impact of selfie filters on FR performance is assessed both in the direct face comparison but also following an occlusion recovery approach, using inpainting techniques to treat the occluded face as an image repairing problem.




\subsection{Occlusion Detection and Segmentation}
For an effective occlusion removal, it is important to accurately detect and segment the occluded facial regions. In the past year, various deep learning-based object detection and segmentation methods have been shown to obtain outstanding performance \cite{8825470,9356353}.

Regions with convolutional neural networks (R-CNNs) \cite{rcnn1}, Fast R-CNNs \cite{rcnn2}, or Faster R-CNNs \cite{rcnn3} are well-known for their state-of-the-art object detection performance. These approaches use selective search algorithms to extract regions from an image, feeding them to a CNN to produce a feature vector for each processed region. Subsequently, machine learning-based classifiers, \eg support vector machines (SVMs), analyse the features extracted from each candidate region to determine the presence of the object. Despite their competitive detection performance, such approaches are computationally expensive. 

The fully convolutional network (FCN) auto-encoder framework \cite{fcn1} is a well performing approach for training an image segmentation network. The need for more robust object boundaries led to the development of U-Net \cite{ronneberger2015unet}, being one of the most used end-to-end FCNs in image segmentation. The U-Net encoder uses a series of convolutions with max pooling layers, while the decoder uses transformed convolutions to upsample the encoded information. The encoder and decoder feature maps are concatenated to better learn the contextual information. For an accurate selfie filter segmentation, the current work adopts the idea presented in \cite{din2020gan}, using U-Net architecture supplemented with a squeeze and excitation (SE) \cite{SENet} block at the output of the first three layers of the encoder.

\subsection{Occlusion Removal}
Deep learning-based algorithms have been effectively used to reconstruct occluded facial regions, \eg \cite{REC3}. On the other hand, inpainting techniques focus on reconstructing the occluded elements of the image, leaving FR out of consideration. It is a challenging task to recover details of facial features on high-level image semantics, being used in many FR scenarios, such as when a subject wears sunglasses \cite{eyeglassRemoval}, a facial mask \cite{din2020gan}, or when there are other external facial occlusions  \cite{khan2019mic,yu2019gate}. The purpose of inpainting is to reconstruct missing information in an image.

Inpainting methods usually consider information from the whole image (\ie low-level texture information and high-level semantic information). Traditional inpainting methods rely on low level information to find best corresponding patches from the unaltered regions in the same image \cite{bertalmio2003in} \cite{barnes2009patch}. These methods work well for background completions and repetitive texture patterns. However, as the face image consists of many unique components, low level features are limited for face inpainting tasks. Thus, the inpainting process needs to be carried out with a high semantic confidence \cite{chen2019gan}.

Facial inpainting (also referred to as face completion) methods have been found to improve FR performance on occluded face images \cite{8987388}.  Rapid progress in deep learning, in particular GANs, inspired lots of studies \cite{chen2018occ} \cite{din2020gan} on facial inpainting. Here, GANs are proposed to deal with both low-level textural features and high-level semantic features utilised for removing facial occlusions. In \cite{din2020gan} several GAN-based image inpainting models, \ie \cite{iizuka2017completion,yu2018context,nazeri2019edge,khan2019mic}, are benchmarked on real world images  showcasing significant reconstruction capability.



\section{Fun Selfie Filter Database}\label{sec:db}
To create the fun selfie filter database used in this work, a qualitative assessment of popular mobile applications for adding fun selfie filters were conducted. In addition, various styles that focus on occluding different facial regions was considered. Subsequently, the selected selfie filters were applied to 1,441 face images of the FRGCv2 \cite{Phillips-FRGC-2005} dataset. For this purpose, an automated software that emulates the chosen mobile applications was used, as illustrated in Fig.~\ref{fig:selfie_filter_creation}. The used subset of the FRGCv2 dataset has good-quality face images which allows to analyse the sole impact of fun selfie filters on FR modules in the absence of quality-related factors \cite{schlett2021face}, \eg variations in pose or illumination.

\subsection{Fun Selfie Filter Selection}
To create an appropriate database of facial images with selfie filters, a total of ten selfie filters were selected from five different fun selfie filter mobile applications. The mobile applications were selected by performing a ranking based on the criteria in Tab.~\ref{self_c}. The scores have been assigned based on the available reviews from users, as well as the authors' experience while using the applications. Tab.~\ref{self_c1} shows the five selfie filter mobile applications that received the highest rankings and which are the mobile applications used in this work. 

\begin{table}[htbp!]
\centering
\caption{Ranking criteria for fun selfie filter applications.}
\label{self_c}
\begin{adjustbox}{max width=\columnwidth}
 \begin{tabular}{@{\extracolsep{2pt}}llll@{}} \toprule 
 & \multicolumn{3}{c}{\textbf{Score}}  \\ \cmidrule{2-4} 
\multirow{-2}{*}{\textbf{Criteria}}  & \multicolumn{1}{c}{\textbf{1}} & \multicolumn{1}{c}{\textbf{2}} & \multicolumn{1}{c}{\textbf{3}} \\ \midrule
Gallery support                    & black-box              & possible               & supported              \\ \midrule
Ease of use                        & hard                   & medium                 & easy                   \\ \midrule
Daily usability                    & no                     & relative               & yes                    \\ \midrule
Variety                            & no                     & relative               & yes                    \\ \midrule
Popularity                         & low                    & medium                 & high                   \\ \midrule
Development                        & no                     & relative               & yes                    \\ \midrule
Complexity                         & easy                   & medium                 & diverse                \\ \midrule
Cost                               & high                   & medium                 & low                    \\ \bottomrule
\end{tabular}
\end{adjustbox}
\end{table}

\begin{table}[htbp!]
\centering
\caption{Selected fun selfie filter applications with the number of downloads and score based on the user reviews.}
\label{self_c1}
\begin{adjustbox}{max width=\columnwidth}
   \begin{tabular}{@{\extracolsep{0pt}}lll@{}} \toprule 
\multicolumn{1}{l}{\textbf{Mobile Application}} & \multicolumn{2}{l}{\textbf{Popularity}} \\ \midrule
Sweet Face Camera  \cite{sel1}                                         & 500M+ downloads & 4.3/5.0 score           \\ \midrule
B612 \cite{sel2}                                                           & 100M+ downloads & 4.5/5.0 score           \\ \midrule
Snow    \cite{sel3}                                                        & 100M+ downloads & 4.2/5.0 score           \\ \midrule
YouCam Fun   \cite{sel4}                                                   & 10M+ downloads & 4.5/5.0 score            \\ \midrule
Bloom Camera   \cite{sel5}                                                        & 1M+ downloads & 4.3/5.0 score             \\ \bottomrule
\end{tabular}
\end{adjustbox}
\end{table}

When investigating the impact of fun selfie filters on FR systems, it is interesting to see how the selfie filter coverage and placement affect the performance of the tested systems. The chosen selfie filters are depicted as part of Fig.~\ref{fig:selfie_filter_creation}.

\subsection{Categorisation}
According to the criteria presented in Tab. \ref{self_c}, a categorisation of fun selfie filters was performed based on facial coverage and placement of the selfie filter.\\

\begin{figure}[h!]
\centering
\includegraphics[width=0.55\linewidth]{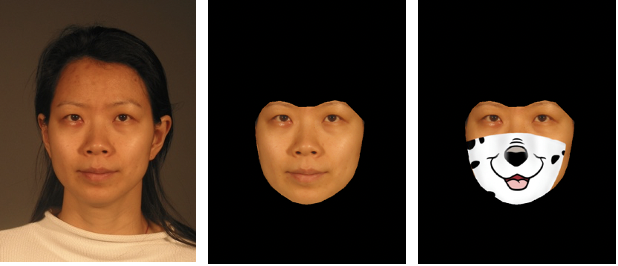}
\caption{Cropping of facial region polygon.}
\label{fac_cov}
\vspace{-15pt}
\end{figure}

\noindent \textbf{Coverage:} The selfie filter coverage is quantified by focusing on the facial region polygon that is used as a mask to the original image, cropping the facial part of the image as illustrated in Fig.~\ref{fac_cov}. 

This information is used further on to investigate the impact of the selfie filter based on its facial alteration, as well as focusing on specific elements that drive the eventual decrease in facial recognition performance.

\begin{equation}
    Coverage_{intensity} = \frac{\Delta \text{Facial Pixel Intensity}}{\text{Number of Facial Pixels}}
    \label{eq:coverage_score}
    \vspace{7pt}
\end{equation}

Eq. \ref{eq:coverage_score} reports on the average pixel intensity variation due to the selfie filter, being a stable and accurate way of estimating the significance of the selfie filter. Using this metric, transparent selfie filters will achieve a lower score in comparison to corresponding solid color selfie filters that cover the same area. Additionally, smoothing, compression, and other effects which do not occlude facial attributes, will not have a big impact on the calculated coverage intensity score. After trying a wide variety of selfie filters provided by various mobile application, the visual complexity of recognising the identity behind the selfie filter has been defined following the thresholds presented in Tab. \ref{sel_gr}. 

\begin{table}[h!]
\caption{Selfie filter groups based on the facial coverage metric.}
\label{sel_gr}
\centering
\begin{adjustbox}{max width=\columnwidth}
\begin{tabular}{@{\extracolsep{2pt}}ll@{}} \toprule 
\textbf{Scenario} & \textbf{$Coverage_{intensity}$}         \\ \midrule
Low Coverage           & less than 15\% difference        \\ \midrule
Medium Coverage        & between 15\% and 40\% difference \\ \midrule
High Coverage          & more than 40\% difference        \\ \bottomrule
\end{tabular}
\end{adjustbox}{}
\end{table}

\begin{figure}[h!]
\centering
\begin{subfigure}[t]{1\columnwidth}
    \centering
  \includegraphics[width=\columnwidth]{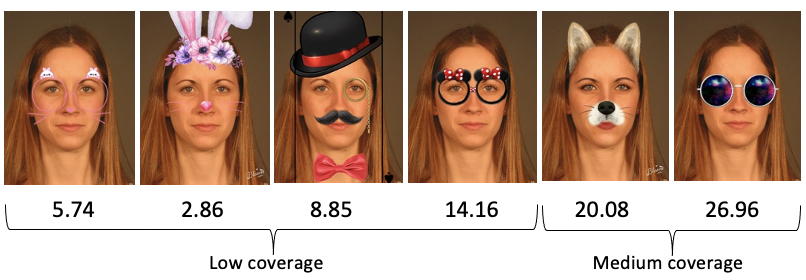}
\end{subfigure}\quad %
\begin{subfigure}[t]{1\columnwidth}
    \centering
  \includegraphics[width=\columnwidth]{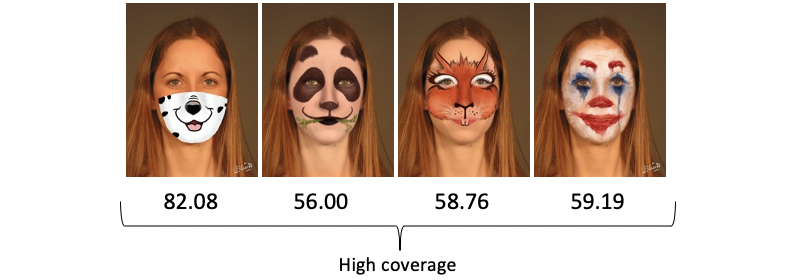}
\end{subfigure}\quad %
\caption{Selfie filters with facial coverage score (see Eq.~\ref{eq:coverage_score})}
\label{coverage_per}
\end{figure}

It is expected that the coverage score  will correlate with the actual difficulty of recognising the original face once the selfie filter is applied. Tab.~\ref{sel_gr} highlights the main selfie filter groups and Fig.~\ref{coverage_per} presents the scores for each of the ten samples. 

\begin{figure}[h!]
\centering
\includegraphics[width=0.9\linewidth]{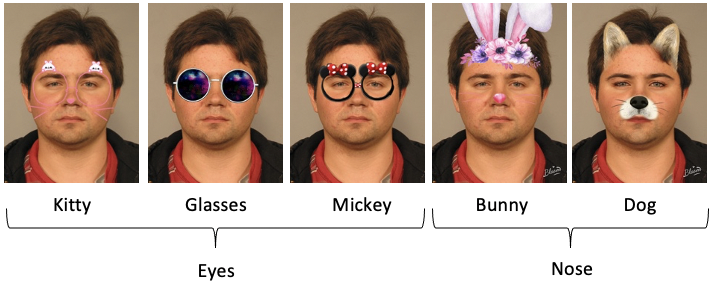}
\includegraphics[width=0.9\linewidth]{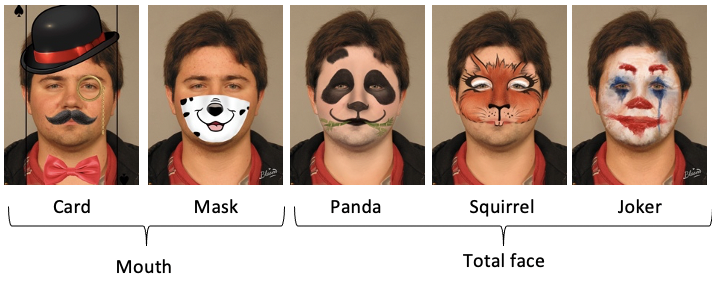}
\caption{Selfie filter groups by the most affected facial region.}
\label{coverage_reg}
\end{figure}

\noindent \textbf{Placement:} Each of the selfie filters presented in Fig.~\ref{coverage_reg} focuses on altering certain facial features. This information will be used in concluding whether certain regions of the face are more important for recognition purposes and if there is any difference across the state-of-the-art FR systems.

\section{Impact on Face Recognition}\label{sec:impact}
The impact of fun selfie filters on face detection and sample quality is estimated for scenarios with different facial coverage measures. In experiments on recognition performance, the most relevant scenario where either one of the face images to be compared has been modified using a fun selfie filter is considered.  For evaluations on recognition performance, the placement of fun selfie filters is additionally considered. In all evaluations a comparison with a baseline of unaltered face images is made.

Fig. \ref{agg_cov1} presents the min-max normalized face detection score, while Tab.~\ref{agg_cov1_t} refers to the actual score, where the detection score ranges differ across the used algorithms (\eg in our case, the range for detection scores on dlib is $[0; 4]$, RetinaFace is $[0; 1]$, and COTS is $[0; 5.4]$).

The min-max normalization is done to ensure an equal scale when comparing the performance of various algorithms as defined in Eq.~\ref{eq:minmaxnorm}.
\begin{equation}
    X_{i}^{\text{normalized}} = \frac{X_{i} - X_{max}}{X_{max} - X_{min}} \cdot (R_{max} - R_{min}) + R_{min}
    \label{eq:minmaxnorm}
\end{equation}
where $R_{min}$ and $R_{max}$ cover the desired range of normalized data (\ie in this case $R_{min} = 0$ and $R_{max} = 1$) and $X_{i}$ refers to the detection score of sample $i$.

\subsection{Face Detection}
\begin{figure}[h!]
    \centering
    \includegraphics[width=1\linewidth]{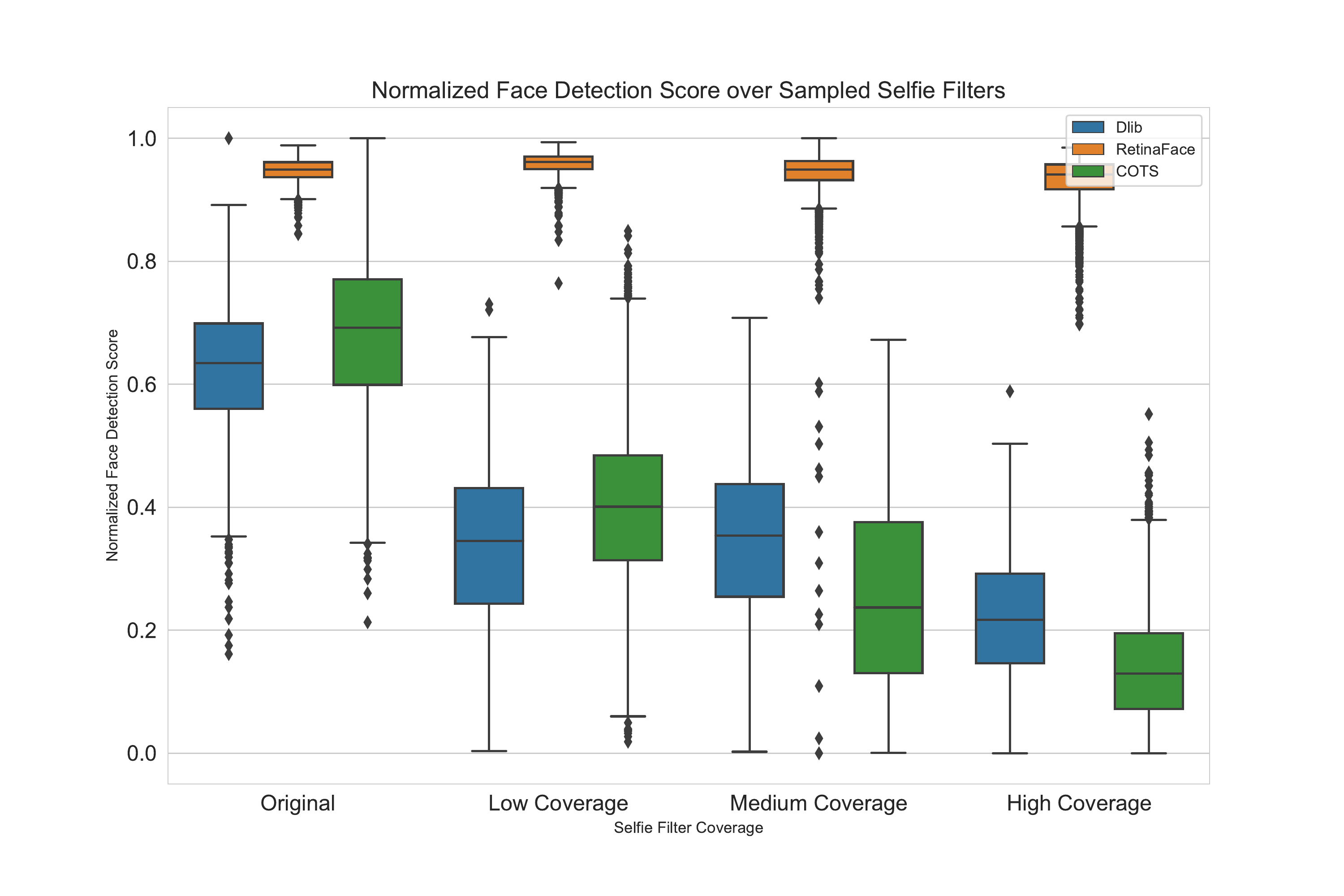}
    \caption{Min-max normalized (eq. \ref{eq:minmaxnorm}) face detection confidence scores by the selfie filter facial coverage.}
    \label{agg_cov1}
    \vspace{-12pt}
\end{figure}

As shown in Tab.~\ref{agg_cov1_t} and Fig.~\ref{agg_cov1}, using dlib \cite{King-MachineLearningToolkit-2009}, RetinaFace \cite{Deng-RetinaFace-CVPR-2020}, and a COTS method, the confidence scores of detected faces for the selfie filtered images, in general, degrade as the selfie filter coverage increases. 

\begin{table}[h!]
\centering
\caption{Average face detection scores, standard deviation, and errors (in \%) by the selfie filter facial coverage.}
\label{agg_cov1_t}
\begin{adjustbox}{max width=\columnwidth}
   \begin{tabular}{@{\extracolsep{2pt}}llllllllll@{}} \toprule 
 & \multicolumn{3}{c}{\textbf{dlib}} & \multicolumn{3}{c}{\textbf{RetinaFace}} & \multicolumn{3}{c}{\textbf{COTS}} \\ \cmidrule{2-4} \cmidrule{5-7} \cmidrule{8-10} 
                              \multirow{-2}{*}{\textbf{Scenario}} & $\bm{\mu}$      & $\bm{\sigma}$      & 
                               $\bm{\epsilon}$ 
                               & $\bm{\mu}$         & $\bm{\sigma}$        & $\bm{\epsilon}$         & $\bm{\mu}$     & $\bm{\sigma}$    & $\bm{\epsilon}$     \\ \midrule
Baseline                          
& 2.031  & 0.529  & 0   &   0.9993  &    0.0002      &     0      &   3.058    &  0.820     &      0     \\ \midrule
Low                          
& 1.343  & 0.462  &  0.681  &      0.9995     &    0.0002      &     0.005      &    1.646   &  0.877    &     0.319      \\ \midrule
Medium                        
& 1.380  & 0.568  &  1.158  &       0.9993    &    0.0005      &     0.015      &   1.265    &  0.964     &     30.55      \\ \midrule
High                           
& 0.874  & 0.388  &  5.658  &       0.9992    &     0.0003     &     2.838      &   0.408    &  0.502     &     21.36      \\ \bottomrule
\end{tabular}
\end{adjustbox}
\end{table}

The used COTS is particularly prone to detection errors ($\epsilon$) when the face is covered at a higher intensity or if the eye region is occluded. Such recognition systems are designed based on constraint environments, where the subject is required to follow some rules for an increased FR pipeline accuracy (\eg in the border control scenario, every person follows a well-defined FR protocol and no unnecessary object is allowed to occlude the face).

\subsection{Sample Quality}
\begin{figure}[h!]
    \centering
    \includegraphics[width=1\linewidth]{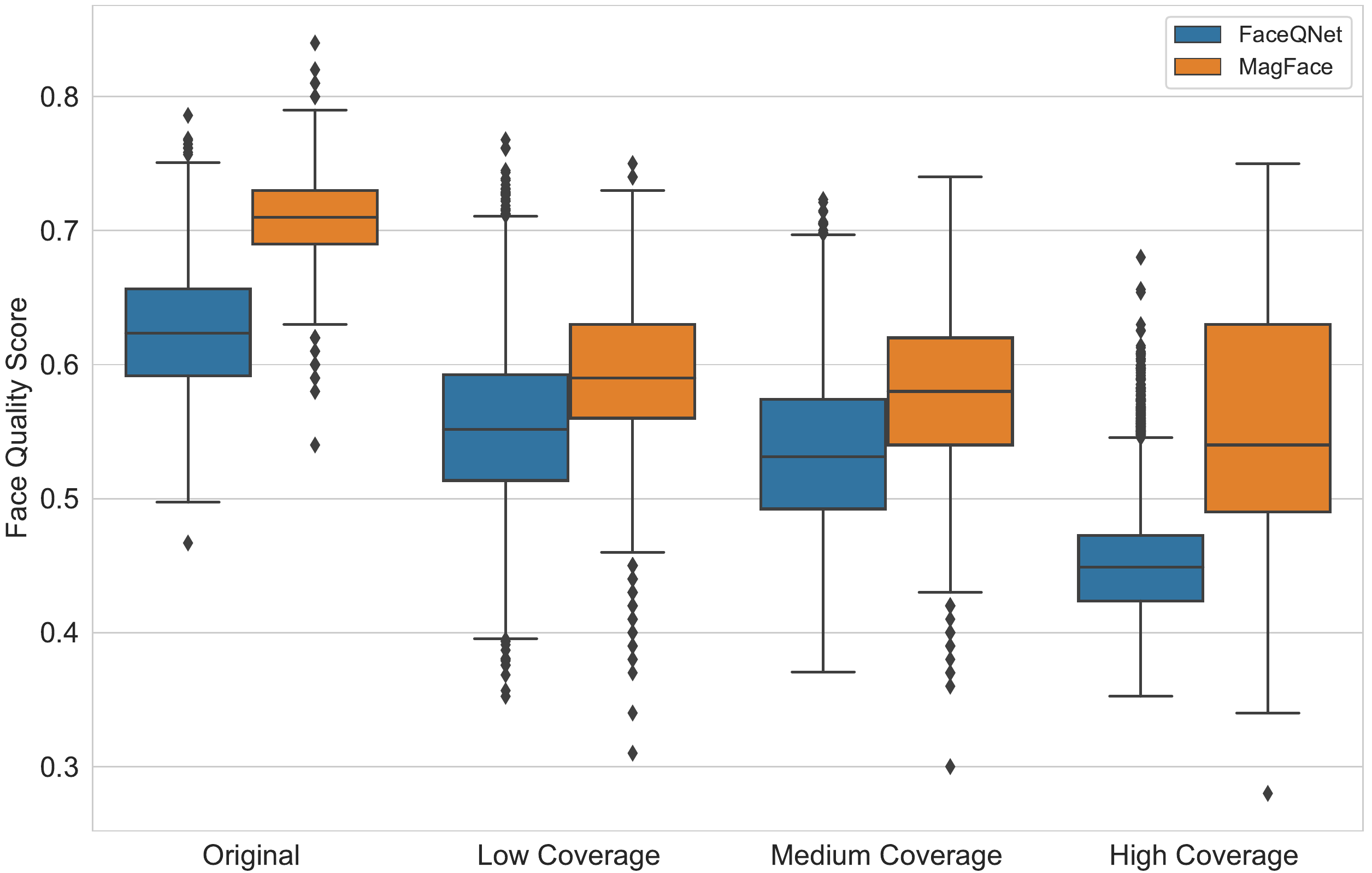}
    \caption{Face quality scores by the selfie filter facial coverage.}
    \label{fqa_cov1}
    \vspace{-11pt}
\end{figure}

For face quality assessment, on the basis of \faceqnet \cite{HernandezOrtega-FQA-FaceQnetV0-ICB-2019} and MagFace \cite{Meng-FRwithFQA-MagFace-IEEE-2021}, results, ranging in the interval $[0; 1]$, are shown in Fig.~\ref{fqa_cov1} and Tab.~\ref{fqa_cov1_tab}.

\begin{table}[h!]
\centering
\caption{Average face quality scores by the selfie filter facial coverage.}
\begin{adjustbox}{max width=\columnwidth}
\begin{tabular}{@{\extracolsep{3pt}}lllll@{}} \toprule 
 & \multicolumn{2}{c}{\textbf{\faceqnet}} & \multicolumn{2}{c}{\textbf{MagFace}}  \\ \cmidrule{2-3} \cmidrule{4-5}
                            \multirow{-2}{*}{\textbf{Scenario}}   & $\bm{\mu}$        & $\bm{\sigma}$       &  $\bm{\mu}$       & $\bm{\sigma}$               \\ \midrule
Baseline  & 0.625    & 0.048      &   0.705     &  0.037              \\ \midrule   
Low Coverage                          
& 0.554    & 0.059    &    0.592     &   0.056               \\ \midrule
Medium Coverage                         
& 0.534    & 0.060     &     0.576    &   0.056            \\ \midrule
High Coverage                           
& 0.450    & 0.044    &    0.556     &    0.081            \\ \bottomrule
\end{tabular}
\end{adjustbox}{}
\label{fqa_cov1_tab}
\end{table}

In addition to the case when the face is fully altered by the selfie filter, a significant effect on face sample quality is shown by medium coverage selfie filters. \faceqnet as well as MagFace return a consistently lower image quality score as the selfie filter coverage increases. For MagFace, the comparatively high variance for high coverage selfie filters is mainly caused by attributing relatively good face quality scores to face images where the joker mask is applied. Hence, if the selfie filter highlights certain facial characteristics, the magnitude of the facial embedding increases while the face may still be significantly occluded.

\subsection{Recognition Performance}
\begin{figure}[h!]
\centering
\begin{subfigure}[t]{0.475\columnwidth}
    \centering
  \includegraphics[width=\columnwidth]{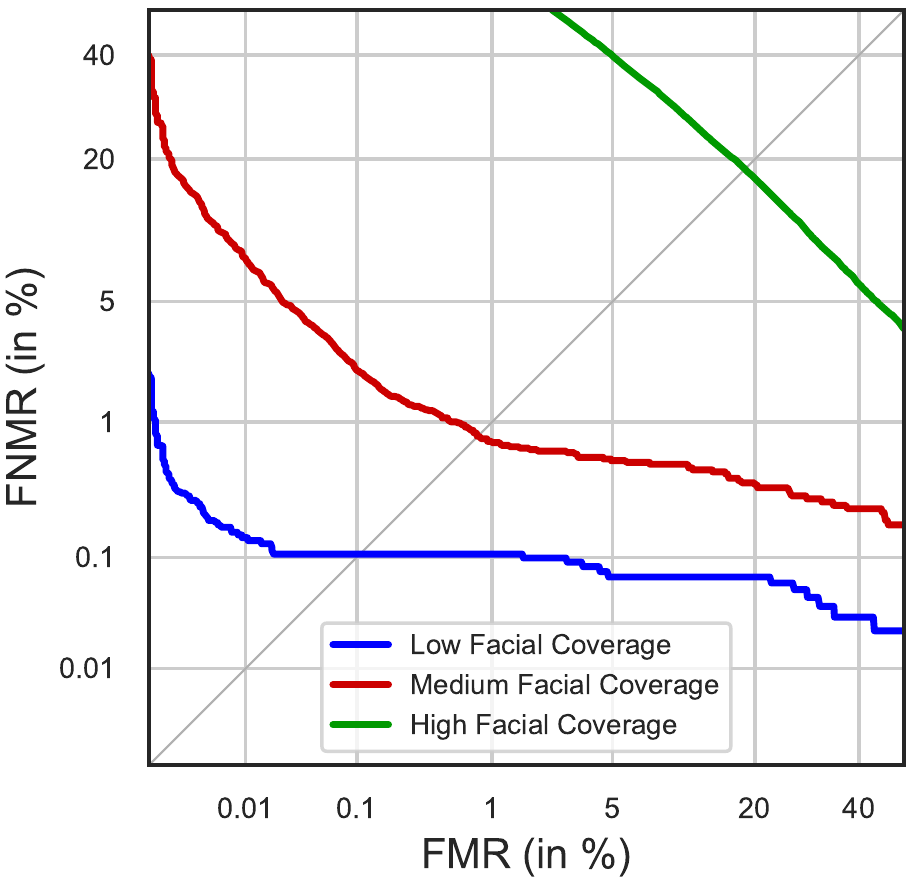}
  \caption{ArcFace}
\end{subfigure}\quad %
\begin{subfigure}[t]{0.475\columnwidth}
    \centering
  \includegraphics[width=\columnwidth]{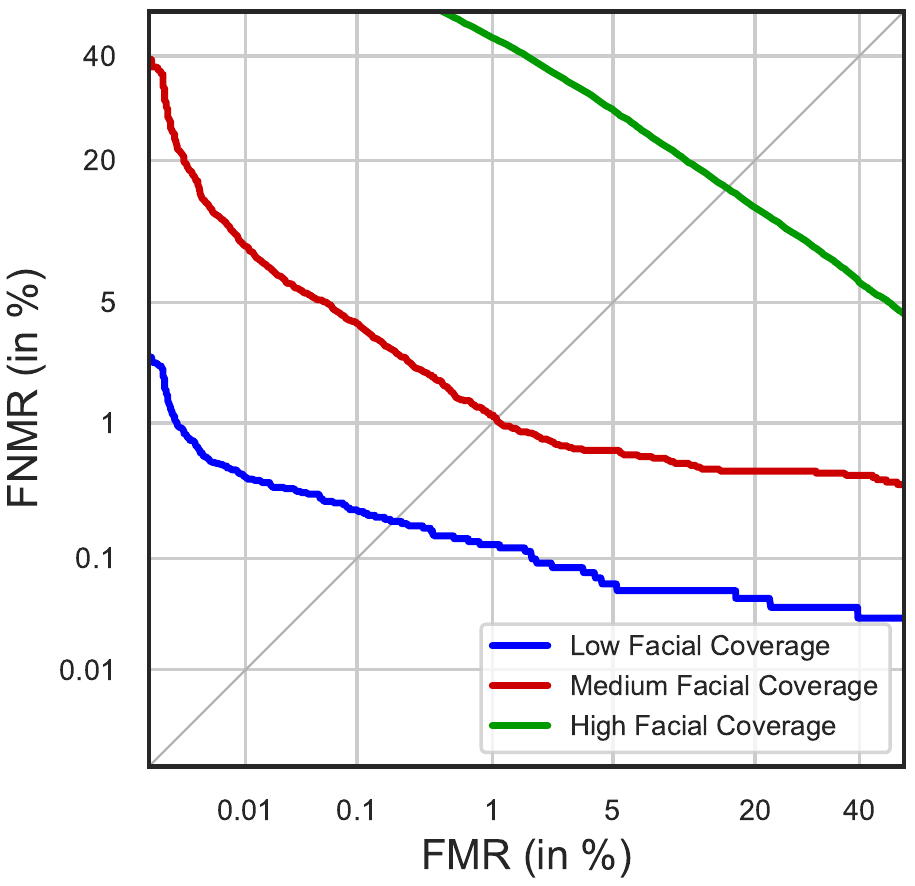}
  \caption{COTS}
\end{subfigure}\quad %
\caption{DET curves by selfie filter coverage.}
\label{_cov_face1}
\end{figure}

Biometric recognition performance is measured in terms of false non-match rate (FNMR) at certain false match rate (FMR) \cite{ISO-IEC-2382-37-121215, ISO-IEC-19795-1-Framework-210216}. In addition, the failure-to-enrol rate (FTE) and the equal error rate (EER) are reported. For the ArcFace \cite{Deng-ArcFace-IEEE-CVPR-2019} and COTS system, a higher selfie filter facial coverage results in a higher FNMR and EER, see detection error trade-off (DET) curves in Fig.~\ref{_cov_face1} and Tab.~\ref{tab:cov_face2_t12}. With respect to the placement of fun selfie filters, the impact of the altered facial region differs but is especially challenging for mouth. Due to its more constraint target environment, COTS performs poorly on FTE when the eyes are covered.

\begin{table}[h!]
\centering
\caption{Face recognition performance (in \%) by fun selfie filter facial coverage.}
\label{tab:cov_face2_t12}
\begin{adjustbox}{max width=\columnwidth}
\begin{tabular}{@{\extracolsep{2pt}}lllllll@{}} \toprule 
\multicolumn{1}{c}{}  &  & \multicolumn{1}{c}{}  & \multicolumn{1}{c}{}  & \multicolumn{3}{c}{\textbf{FNMR}} \\ \cmidrule{5-7} 
{\multirow{-2}{*}{\textbf{System}}} & \multirow{-2}{*}{\textbf{Scenario}} & \multirow{-2}{*}{\textbf{FTE}} & \multirow{-2}{*}{\textbf{EER}} &  \textbf{FMR$=$0.01}     & \textbf{FMR$=$0.1}      & \textbf{FMR$=$1}      \\ \midrule
\multirow{4}{*}{ArcFace} & Baseline  & 0  & 0  & 0  & 0   & 0            \\
& Low Coverage  & 0.005  & 0.100  & 0.210  & 0.100   & 0.100            \\
& Medium Coverage & 0.015  & 0.850  & 9.400  & 3.500   & 0.830            \\
& High Coverage  & 2.838  & 19.80  & 73.00  & 59.20   & 48.80            \\ \midrule
\multirow{4}{*}{COTS} & Baseline & 0  & 0  & 0  & 0   & 0            \\
& Low Coverage  & 0.319  & 0.250  & 0.650  & 0.400   & 0.175            \\
& Medium Coverage & 30.55  & 1.050  & 10.70  & 4.100   & 1.010            \\
& High Coverage & 21.36  & 18.20  & 66.00  & 50.60   & 44.70            \\ \bottomrule
\end{tabular}
\end{adjustbox}
\end{table}

\begin{figure}[h!]
\centering
\begin{subfigure}[t]{0.475\columnwidth}
    \centering
  \includegraphics[width=\columnwidth]{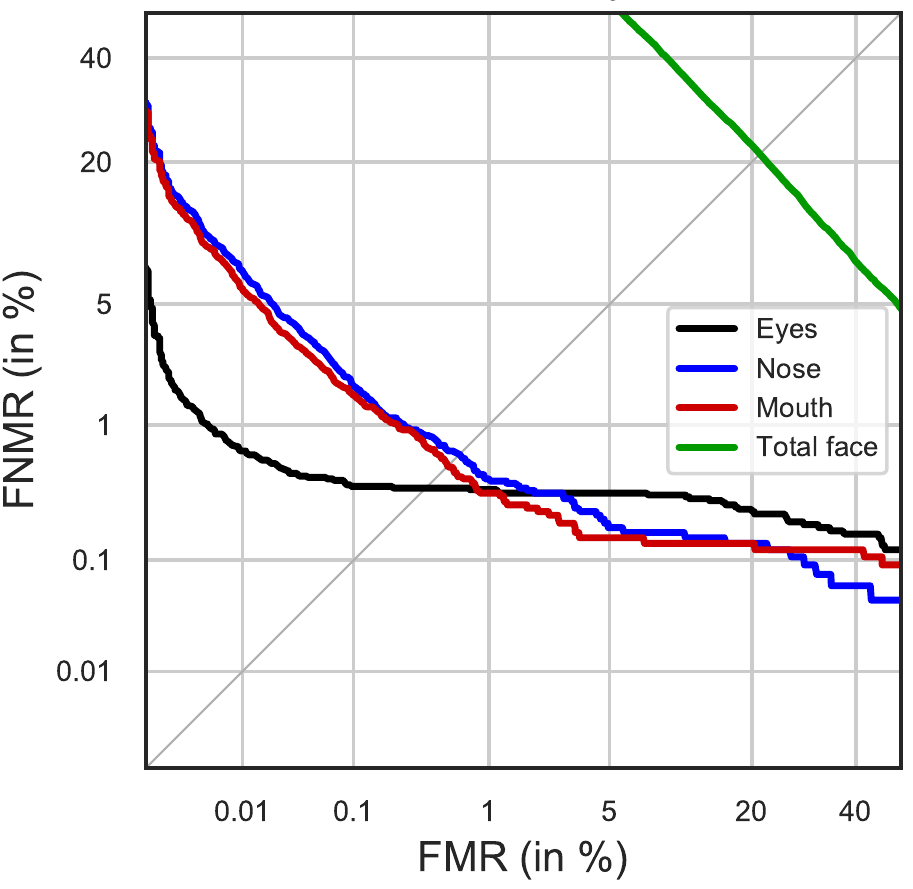}
  \caption{ArcFace}
\end{subfigure}\quad %
\begin{subfigure}[t]{0.475\columnwidth}
    \centering
  \includegraphics[width=\columnwidth]{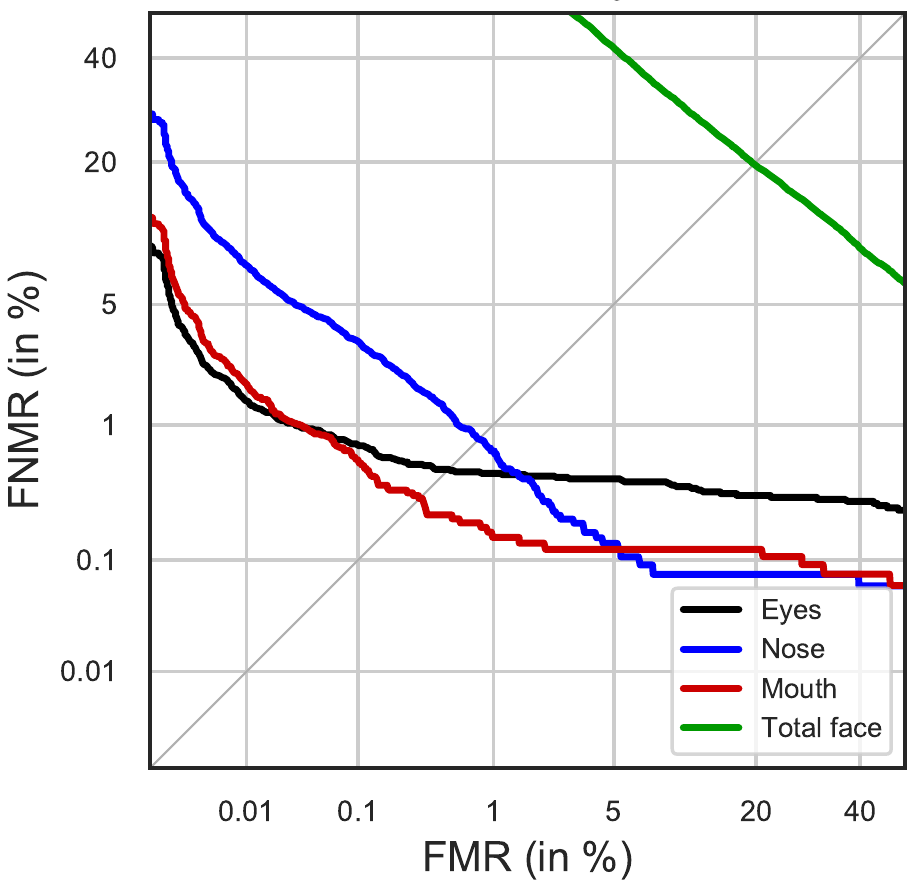}
  \caption{COTS}
\end{subfigure}\quad %
\caption{DET curves by selfie filter placement.}
\label{_cov_face2}
\end{figure}

\begin{table}[h!]
\centering
\caption{Face recognition performance (in \%) by the facial region that is altered by the fun selfie filter.}
\label{tab:pl_face2_t12}
\begin{adjustbox}{max width=\columnwidth}
\begin{tabular}{@{\extracolsep{2pt}}lllllll@{}} \toprule 
\multicolumn{1}{c}{}  &  & \multicolumn{1}{c}{}  & \multicolumn{1}{c}{}  & \multicolumn{3}{c}{\textbf{FNMR}} \\ \cmidrule{5-7} 
\multirow{-2}{*}{\textbf{System}} & \multirow{-2}{*}{\textbf{Scenario}} & \multirow{-2}{*}{\textbf{FTE}} & \multirow{-2}{*}{\textbf{EER}} &  \textbf{FMR$=$0.01}     & \textbf{FMR$=$0.1}      & \textbf{FMR$=$1}      \\ \midrule
\multirow{5}{*}{ArcFace} & Baseline  & 0  & 0  & 0  & 0   & 0 \\
& Mouth & 0.244  & 0.820  & 6.200  & 2.200   & 0.550            \\
& Eyes  & 0  & 0.650  & 0.870  & 0.550   & 0.500            \\ 
& Nose   & 0.139  & 0.850  & 8.800  & 2.780   & 0.620            \\ 
& Total face & 0.418  & 22.90  & 94.80  & 78.60   & 56.80            \\ \midrule
\multirow{5}{*}{COTS} & Baseline & 0  & 0  & 0  & 0   & 0 \\
& Mouth & 0.662  & 0.500  & 2.200  & 0.700   & 0.210            \\
& Eyes  & 20.66  & 0.760  & 1.920  & 0.910   & 0.710            \\
& Nose  & 0.069  & 0.900  & 9.700  & 3.750   & 0.730            \\
& Total face & 28.97  & 19.80  & 82.90  & 69.70   & 58.30            \\ \bottomrule
\end{tabular}
\end{adjustbox}
\end{table}

\begin{figure*}[t]
\centering
\includegraphics[width=0.85\linewidth]{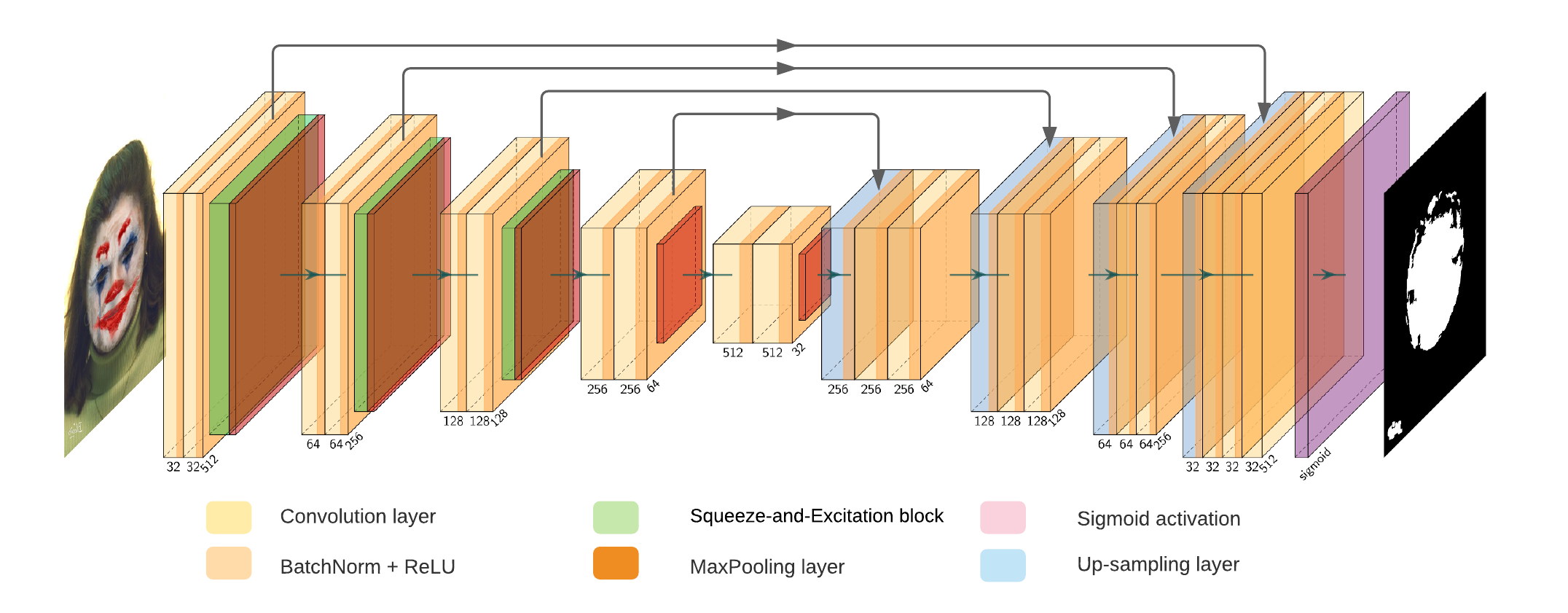}
\caption{Architecture of the fun selfie filter segmentation network.}
\label{segmentation}
\vspace{-10pt}
\end{figure*}
\section{Fun Selfie Filter Removal}
\label{sec:gan}
The facial coverage resulting from the application of fun selfie filters can range from being almost non-existent to the extreme case of full coverage (\cf Fig.~\ref{coverage_per}). As shown in the previous section, FR systems recognize well on low facial coverage scenarios, while they are rather challenged by selfie filters that produce high facial coverage. 



This section introduces the proposed fun selfie filter removal method which represents an adaptation of the inpainting technique of \cite{yu2019gate}. As suggested in \cite{din2020gan}, the architecture is supplemented with a perceptual network \cite{simonyan2015nets}, in a form of a pre-trained VGG-19 fixed network, to stimulate the generator output, to have similar feature representation to the ground truth ones.
The segmentation module and perceptual network are inspired from \cite{chen2019gan}. As proposed in \cite{chen2019gan}, a good way of improving the GAN-based image inpainting accuracy is to split the occluded image segmentation from the actual image inpainting. 

\subsection{Segmentation}
An overview of the architecture of the segmentation algorithm is depicted in Fig.~\ref{segmentation}. The output of the segmentation module is a binary map indicating the pixels covered by the selfie filter. The generator of the segmentation map is a modified version of the U-Net architecture~\cite{ronneberger2015unet} consisting of a CNN-based encoder and decoder: 

\begin{itemize}[leftmargin=*]
    \item[] \textbf{Encoder} consists of five blocks comprising of a convolution and a 'Squeeze-and-Excitation' layer, followed by a down-sampling of the input along its spatial dimensions by applying a \textit{MaxPool} of \textit{kernel size} 2 and \textit{stride} 2;
    \item[] \textbf{Decoder} resembles the encoder architecture except that the \textit{MaxPool} is replaced by the up-sampling layer and instead of the convolution layers, deconvolution layers are applied, where the last layer of the decoder uses \textit{sigmoid} activation function.
\end{itemize}

The local information is combined with the global one by concatenating the result of the deconvolution layers with the feature maps from the encoder at the same level. As a loss function, cross-entropy is used between the predicted binary map and corresponding target map, adding a post processing step to handle image processing operations of erosion and dilation.

\subsection{Inpainting}
The goal of this module is to remove the selfie filter and reconstruct the representation of the facial characteristics that have been covered by the selfie filter in a way that is both structural and appearance wise consistent with the ground truth image. The main building blocks for the image inpainting module are the generator, the discriminator and the perceptual network.

\textbf{Generator:}
\label{generator}
The generator has the same encoder and decoder architecture as the generator of the segmentation map, with the addition of gated convolution for the image inpainting network, accounting for a dynamic feature selection mechanism for each channel and spacial location.

\begin{figure*}[h!]
\centering
\includegraphics[width=\linewidth]{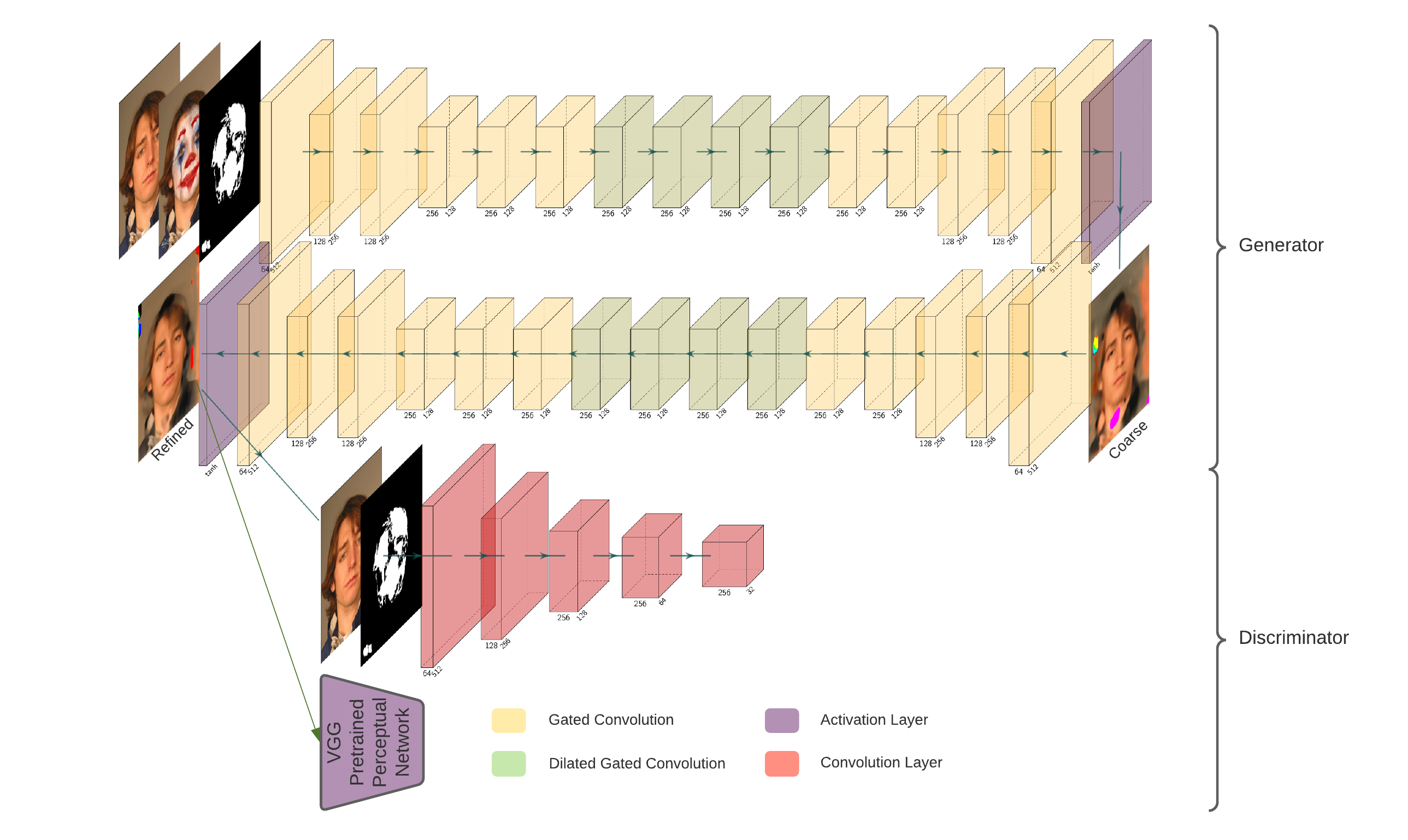}
\caption{Architecture of the inpainting network.}
\label{editor_gan}
\vspace{-10pt}
\end{figure*}

Fig. \ref{editor_gan} depicts the used architecture, where each convolution is distinctively marked based on its type (\eg gated, dilated gated, or normal convolution). Overall, the GAN model takes as input the original or selfie filtered image together with the corresponding selfie filter binary segmentation and passes it to the first generator network. Once the coarse output is derived, it is passed through a refinement network for an improved inpainting. The refined inpainting together with the selfie filter binary segmentation is the input to the fully convolutional discriminator and to the perceptual network. As a result, the GAN loss is computed and the training proceeds until it has reached the targeted number of iterations.

\textbf{Discriminator}
\label{discriminator}
For training free-form image inpainting networks, the fully convolutional discriminator architecture is used. As indicated earlier, the network is inspired by global and local GANs \cite{iizuka2017completion}, MarkovianGANs \cite{isola2017adversarial}, and perceptual loss \cite{johnson2016perc}.

A six strided convolutional network with kernel size 5 and stride 2 is used as the discriminator. Additionally, GANs are applied for each feature element in this feature map, formulating $h \times w \times c$ number of GANs focusing on different locations and different semantics of the input image.

\textbf{Perceptual network}
The third module of the image inpainting module presented in Fig.~\ref{editor_gan} is a perceptual network, in the form of a pre-trained VGG-19 fixed network \cite{simonyan2015nets} with a perceptual loss \cite{johnson2016perc} that is applied to penalize the outputs that are perceptually not reasonable by defining a feature level distance measure between the intermediate feature maps of the reconstructed image and its original counterpart. The purpose of this network is to encourage the generator's output to be similar to the original image.

As a optimization, \cite{din2020gan} suggests exploiting the intermediate convolution layer feature maps of the VGG-19 network to get rich structural information. This is expected to help in recovering plausible structure of the face semantics. 

The overall generator loss function, $\mathcal{L}$, is defined as:
\begin{equation}
    \mathcal{L}= \lambda_{\text{rc}_{\text{coarse}}} \cdot \mathcal{L}_{\text{rc}_{\text{coarse}}} + \lambda_{\text{rc}_{\text{refined}}} \cdot  \mathcal{L}_{\text{rc}_{\text{refined}}} + \lambda_{\text{perc}} \cdot \mathcal{L}_{\text{perc}} + \lambda_{G} \cdot \mathcal{L}_{G}
\end{equation}

\noindent where $\lambda_{\text{rc}_{\text{coarse}}} = 30$, $\lambda_{\text{rc}_{\text{refined}}} = 70$, $\lambda_{\text{perc}} = 50$, and $\lambda_{G} = 0.7$.

$\mathcal{L}_{\text{rc}} = \mathcal{L}_{\text{H}} + \mathcal{L}_{\text{SSIM}}$ (calculated based on the coarse and refined outputs, as presented in Fig.~\ref{editor_gan}). $\mathcal{L}_{\text{H}}$ uses the mean squared error (MSE) if the absolute element-wise error falls below one and the $\textit{l}_{\text{1}}$-distance, otherwise. Its combination with the $\mathcal{L}_{\text{SSIM}}$ ensures that the resulting image resembles its target, being also similar in terms of structural similarity index (SSIM). $\mathcal{L}_{\text{perc}}$ refers to the perceptual loss, while $\mathcal{L}_{G}$, the generator loss, captures the MSE loss given the discriminator's refined output and the target image.

Additionally, the discriminator loss function, $ \mathcal{L_{D}}$, is:
\begin{equation}
    \mathcal{L_{D}} = 0.5 \cdot (\mathcal{L_{\text{MSE}_{\text{fake}}}} + \mathcal{L_{\text{MSE}_{\text{real}}}})
\end{equation}

\noindent where $\mathcal{L_{\text{MSE}}}$ is the MSE between the input and target tensor.

For both the generator and the discriminator, the Adam optimizer is applied \cite{adam-optim} with an initial learning rate of 0.001 that is adjusted every 50,000 training iterations by 0.1.
\subsection{Training}
\begin{figure}[h!]
\centering
\includegraphics[scale = 0.38]{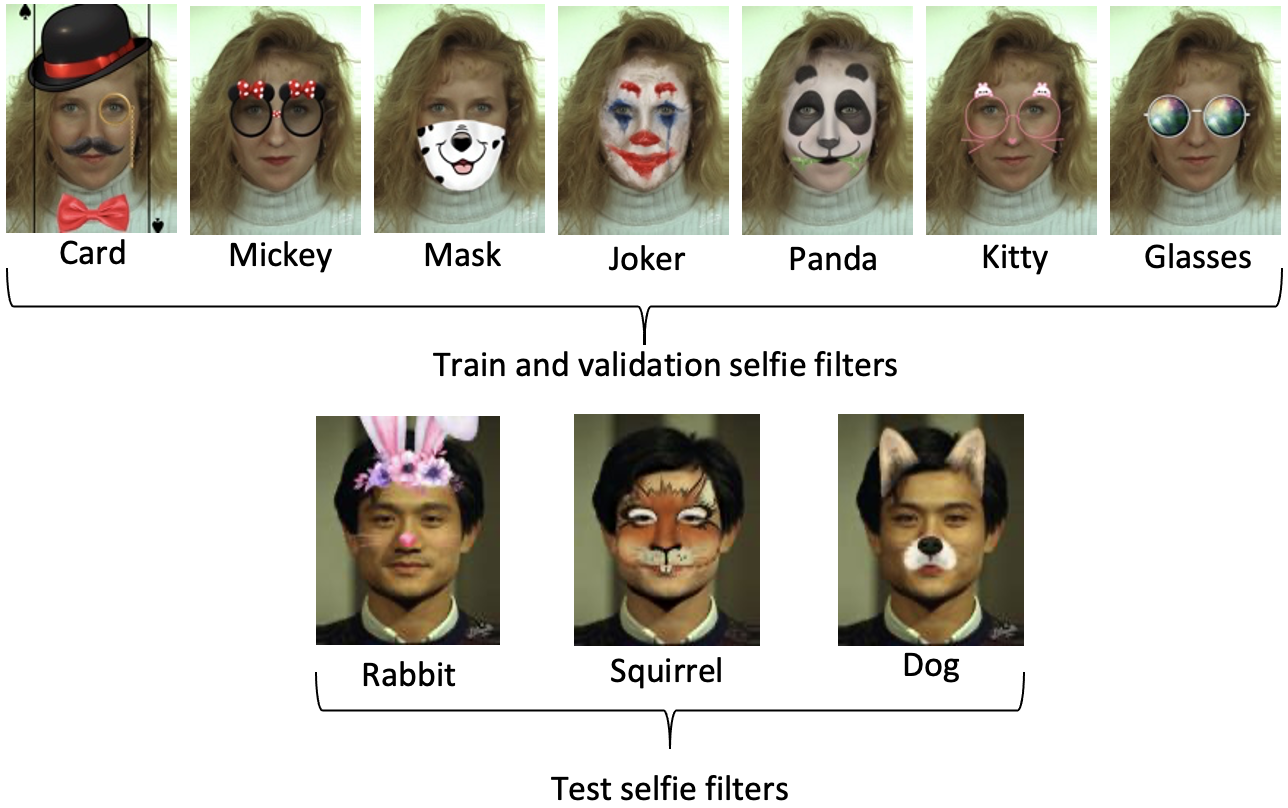}
\caption{Selfie filters used for training, validation and testing.}
\label{tt}
\end{figure}

To assess the GAN model's generalizability, seven selfie filters are chosen for training and validation, while the remaining three are used for testing, as shown in Fig. \ref{tt}.

The overall image inpainting process is directly related to the quality of the selfie filter segmentation. Hence, for enhanced accuracy and generalizability, the training of the segmentation module should be done on a wide variety of selfie filters. Given the limited number of considered popular selfie filters (Fig. \ref{tt}), it becomes rather challenging to generalize well on unknown selfie filter occluding unknown facial images. Therefore, data augmentation was applied.

\begin{figure}[h!]
\centering
\includegraphics[width=1\columnwidth]{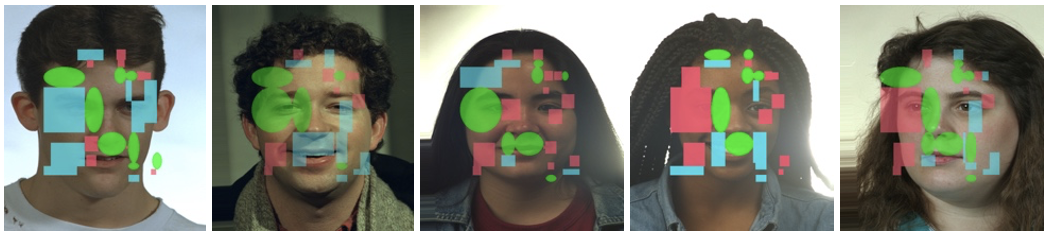}
\caption{Data augmentation: Overlay with multiple shapes of random color, size, and transparency.}
\label{data_aug_par}
\end{figure}

The implemented data augmentation method, for which examples are shown in Fig. \ref{data_aug_par}, consists of three steps:
\begin{enumerate}
    \item Identify the facial region based on its landmarks;
    \item Divide the identified region in a the desired number of subregions that do not overlap;
    \item Place random shapes (of random colour and intensity) on a subset of subregions such that only one shape is attributed to a subregion and its size is a perfect fit for the target subregion.
\end{enumerate}

In a cross-database experiment, the training is performed on the FERET dataset, while the selfie filter removal is applied to the FRGC dataset.

The training of the GAN based model for selfie filter removal was done on a Tesla M10 GPU over 12,225 training samples of size 512 $\times$ 512, where 4,355 are selfie filtered images (Fig. \ref{tt}) based on the FERET references and 7,870 are semi-synthetically created images with shapes (Fig. \ref{data_aug_par}) based on the FERET probes. The model is trained for 70 epochs (i.e., as a good time and performance trade-off), using an Adam optimizer as a replacement optimization algorithm for stochastic gradient descent.

\begin{figure}[h!]
\centering
\includegraphics[scale = 0.7]{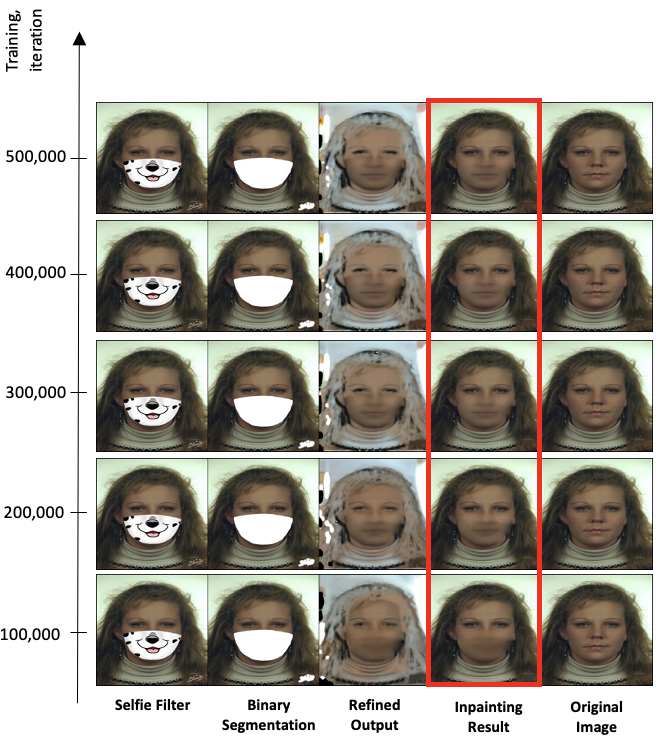}
\caption{Incremental GAN based Selfie Filter removal training.}
\label{train_gan}
\end{figure}

\begin{figure}[h!]
\centering
\includegraphics[scale = 0.5]{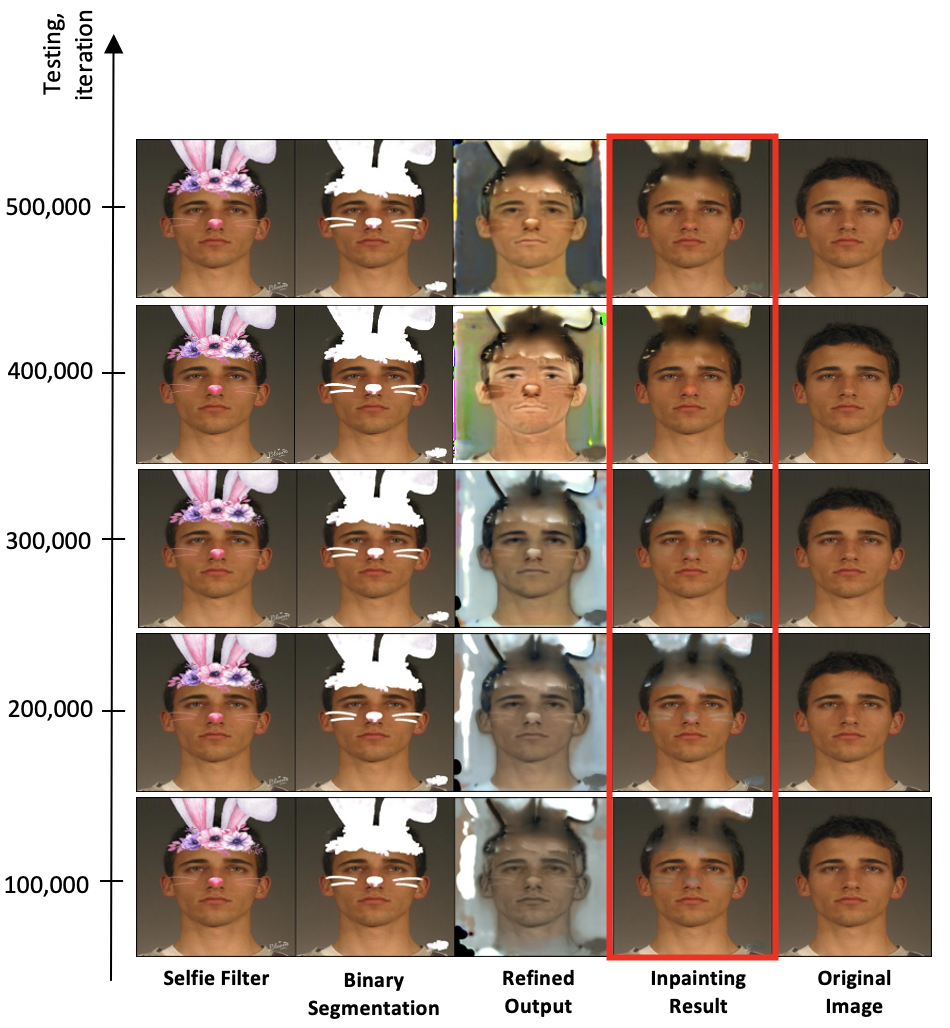}
\caption{GAN testing on removing unseen Selfie Filters.}
\label{train_gan2}
\end{figure}

Fig. \ref{train_gan} presents the evolution of the GAN based selfie filter removal training, where samples exceeding 500,000 iterations do not differ much from the performance attained just after 500,000 iterations. However, fine-tuning the generator and the discriminator over more iterations was seen to be very important when testing the selfie filter removal. Fig.~\ref{train_gan2} highlights the selfie filter removal performance on unseen selfie filters over a set of pre-trained weights.

\subsection{FR Performance with Selfie Filter Removal}

\label{sec:results}

Having trained the GAN-based selfie filter removal model for at least 500,000 iterations, following the train and test split presented in Fig. \ref{tt}, a comparison with the original unaltered counterparts is performed. 

Following the Peak Signal-to-Noise Ratio (PSNR) \cite{psnr} and Mean Structural Similarity Index (MSSIM) \cite{ssim} scores, Tab.~\ref{score_rem} indicates an overall higher similarity between the FRGC reference and the image where the selfie filter has been removed relative to the selfie filtered counterpart.

\begin{table}[h!]
\centering
\caption{The PSNR and MSSIM scores for the selfie filtered image (Fig.~\ref{tt}) and its reconstructed counterpart, averaged over 1,441 images.}
\label{score_rem}
\begin{adjustbox}{max width=\columnwidth}
\begin{tabular}{@{\extracolsep{2pt}}lllcccc@{}} \toprule 
\multicolumn{1}{c}{}  &  & \multicolumn{1}{c}{}  & \multicolumn{2}{c}{\textbf{Selfie Filtered}} & \multicolumn{2}{c}{\textbf{\begin{tabular}[c]{@{}c@{}}Reconstructed Image\\ (Selfie filter removal)\end{tabular}}} \\ 
\cmidrule{4-5} 
\cmidrule{6-7} 
{\multirow{-2}{*}{\textbf{Selfie filter}}} & \multirow{-2}{*}{\textbf{Coverage}} &
\multirow{-2}{*}{\textbf{Placement}} &  \textbf{PSNR}     & \textbf{MSSIM}      & \textbf{PSNR}   & \textbf{MSSIM}   \\ \midrule
Card & Low  & mouth & 22.42 & 0.967 & \textbf{25.32} & \textbf{0.970} \\ 
Kitty & Low  & eyes & 22.12 & 0.965 & \textbf{27.88} & \textbf{0.967} \\ 
Bunny & Low  & nose & 21.55 & 0.966 & \textbf{34.71} & \textbf{0.976} \\ 
Mickey & Low  & eyes & 22.08 & 0.966 & \textbf{23.85}  & \textbf{0.968} \\  \midrule
Glasses  & Medium &  eyes & 19.44 & 0.927  & \textbf{22.11} & \textbf{0.968} \\ 
Dog  & Medium  & nose & 21.27 & 0.955 & \textbf{23.96} & \textbf{0.964} \\  \midrule
Mask  & High  & mouth & 16.26 & 0.930 & \textbf{22.42} & \textbf{0.959} \\ 
Panda  & High  & total & 18.87 & 0.915 & \textbf{19.10} & \textbf{0.943} \\ 
Squirrel  & High  & total & 20.60 & 0.920 & \textbf{22.52} & \textbf{0.964} \\ 
Joker  & High  & total & 19.60 & 0.902 & \textbf{26.82} & \textbf{0.945} \\  \bottomrule
\end{tabular}
\end{adjustbox}
\end{table}

\textit{PSNR} captures the ratio between the maximum possible power of a signal and the power of corrupting noise that affects the fidelity of its representation.

\textit{MSSIM} is a perception-based model that considers image degradation as perceived change in structural information, where pixels have strong inter-dependencies especially when they are spatially close. A higher \textit{PSNR} or \textit{MSSIM} score relate to an increased degree of similarity between the compared images.

To quantify the benefit of applying the proposed removal method, the FR performance is re-evaluated after removing the selfie filters. The corresponding DET curves are plotted for selfie filters of various facial coverage (Fig.~\ref{coverage_per}) and affected facial regions (Fig.~\ref{coverage_reg}).

\begin{figure}[h!]
\centering
\begin{subfigure}[t]{0.48\columnwidth}
    \centering
  \includegraphics[width=\columnwidth]{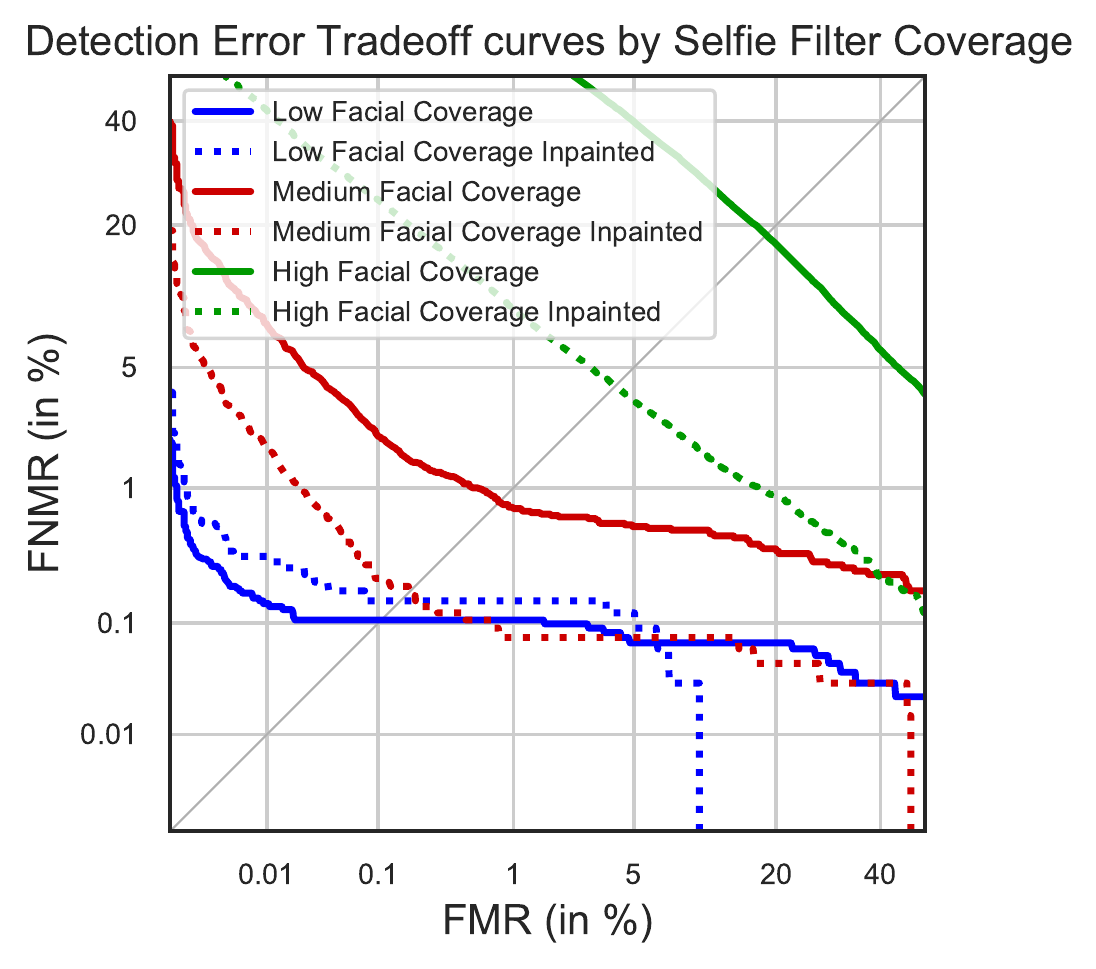}
  \caption{ArcFace}
  \label{_res1}
\end{subfigure}\quad%
\begin{subfigure}[t]{0.48\columnwidth}
    \centering
  \includegraphics[width=\columnwidth]{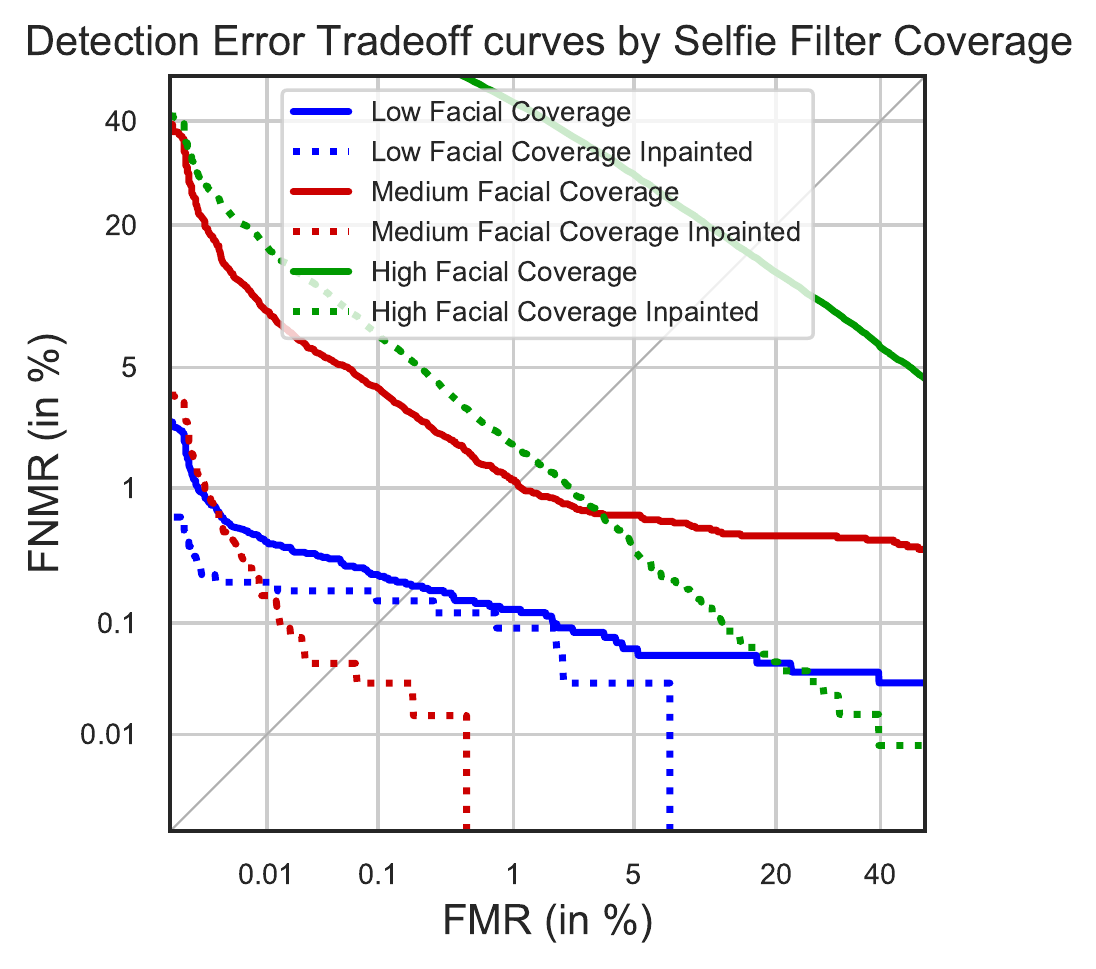}
  \caption{COTS}
  \label{_res2}
\end{subfigure}\quad %
\caption{DET curves for FRGC based selfie filter test dataset by Selfie Filter Coverage.}
\label{_res}
\end{figure}

\begin{table}[h!]
\centering
\caption{EER comparison between the selfie filtered and its selfie filter removal counterpart by Selfie Facial Coverage.}
\label{GAN_result1}
\begin{adjustbox}{max width=\columnwidth}
\begin{tabular}{@{\extracolsep{2pt}}llccc@{}} \toprule 
\multicolumn{1}{c}{}  &  & \multicolumn{1}{c}{\textbf{Selfie Filtered}} & \multicolumn{1}{c}{\textbf{\begin{tabular}[c]{@{}c@{}}Selfie Filter Removal\end{tabular}}} &  \\ 
\multirow{-2}{*}{\textbf{FR System}} & \multirow{-2}{*}{\textbf{Coverage}} &
\textbf{EER \%} &  \textbf{EER \%}  & \multirow{-2}{*}{\textbf{$\Delta$}}  \\ \midrule
ArcFace & Low & 0.100 & 0.200 & 0.100 \\ 
& Medium & 0.850 & 0.250 &\textbf{-0.600} \\ 
& High & 19.80 & 4.500 & \textbf{-15.30} \\ \midrule
COTS & Low & 0.250 & 0.210 & \textbf{-0.040}\\ 
& Medium & 1.050 & 0.060 & \textbf{-0.990} \\ 
& High & 18.20 & 2.100 & \textbf{-16.10} \\  \bottomrule
\end{tabular}
\end{adjustbox}
\end{table}

Comparing the results in Fig.~\ref{_res} and Tab.~\ref{GAN_result1} with their selfie filtered counterparts, the EER has mostly decreased, improving the FR performance across facial coverage and placement. The only exception is proposed by the low coverage selfie filters, where, due to some facial artifacts after the removal, the metric is slightly higher than the selfie filtered counterpart, while still keeping the superior FR performance.

\begin{figure}[h!]
\centering
\begin{subfigure}[t]{0.29\columnwidth}
  \centering
  \includegraphics[width=\linewidth]{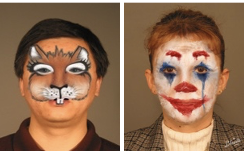}
  \caption{Selfie filtered}
  \label{_dlib_f}
\end{subfigure} \quad %
\begin{subfigure}[t]{0.29\columnwidth}
  \centering
  \includegraphics[width=\linewidth]{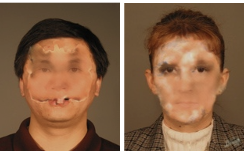}
  \caption{Removal}
  \label{_ret_f}
\end{subfigure} \quad %
\begin{subfigure}[t]{0.29\columnwidth}
  \centering
  \includegraphics[width=\linewidth]{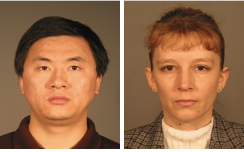}
  \caption{Original}
  \label{_ret_f}
\end{subfigure} \quad %
\caption{Examples of total facial coverage selfie filters.}
\label{_rec_tf}
\vspace{-5pt}
\end{figure}

\begin{figure}[h!]
\centering
\begin{subfigure}[t]{0.29\columnwidth}
  \centering
  \includegraphics[width=\linewidth]{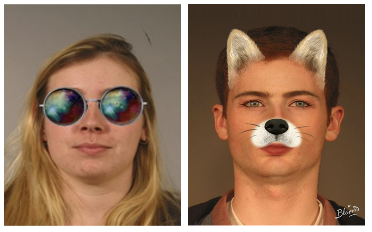}
  \caption{Selfie filtered}
\end{subfigure} \quad %
\begin{subfigure}[t]{0.29\columnwidth}
  \centering
  \includegraphics[width=\linewidth]{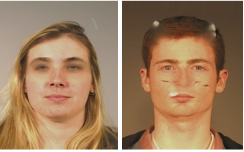}
  \caption{Removal}
\end{subfigure} \quad %
\begin{subfigure}[t]{0.29\columnwidth}
  \centering
  \includegraphics[width=\linewidth]{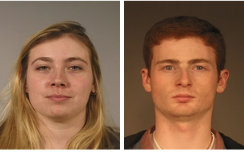}
  \caption{Original}
\end{subfigure} \quad %
\caption{Examples of medium facial coverage selfie filters.}
\vspace{-10pt}
\label{_rec_tf_a}
\end{figure}

The high coverage selfie filters have benefited the most from the proposed selfie filter removal approach, achieving a 15.30\% percentage point lower EER for ArcFace (Fig.~\ref{_res1}) and a 16.10\% percentage point lower EER for COTS (Fig.~\ref{_res2}) compared to the selfie filtered variant (Fig.~\ref{_rec_tf_a}). 

\begin{figure}[h!]
\centering
\begin{subfigure}[t]{0.29\columnwidth}
  \centering
  \includegraphics[width=\linewidth]{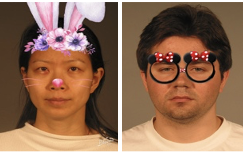}
  \caption{Selfie filtered}
\end{subfigure} \quad %
\begin{subfigure}[t]{0.29\columnwidth}
  \centering
  \includegraphics[width=\linewidth]{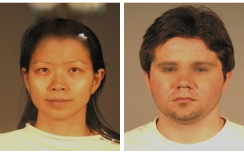}
  \caption{Removal}
\end{subfigure} \quad %
\begin{subfigure}[t]{0.29\columnwidth}
  \centering
  \includegraphics[width=\linewidth]{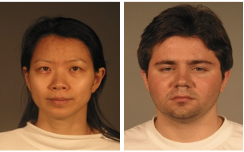}
  \caption{Original}
\end{subfigure} \quad %
\caption{Examples of low facial coverage selfie filters.}
\label{_rec_tf_b}
\vspace{-5pt}
\end{figure}

For medium facial coverage selfie filters, despite of a realistic reconstruction of the face, the challenge of approximating facial elements of the mouth region in particular has not allowed for a significant growth in FR performance (Fig. \ref{_rec_tf_b}).



In the case of low coverage selfie filters, facial images before and after selfie filter removal maintain a high visibility of the original facial characteristics (Fig.~\ref{_rec_tf_b}). 

Despite of the relatively accurate selfie filter removal, the resulting images might vary in brightness or might still contain selfie filter related artefacts. In the case of the total facial coverage only some facial elements are approximated, while still leading to an enhanced FR performance. 


\begin{figure}[h!]
\centering
\begin{subfigure}[t]{0.48\columnwidth}
  \centering
  \includegraphics[width=1\linewidth]{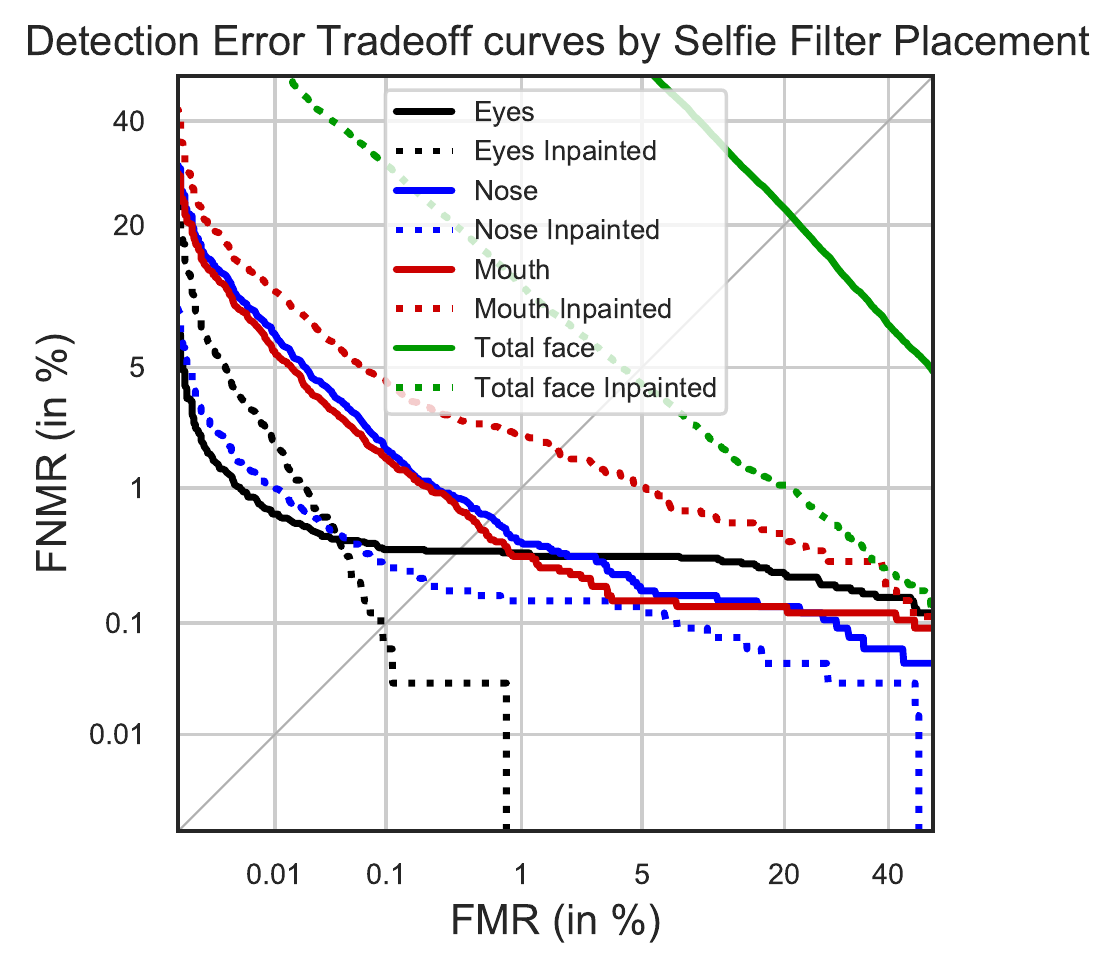}
  \caption{ArcFace}
  \label{_res221}
\end{subfigure}\quad%
\begin{subfigure}[t]{0.48\columnwidth}
  \centering
  \includegraphics[width=1\linewidth]{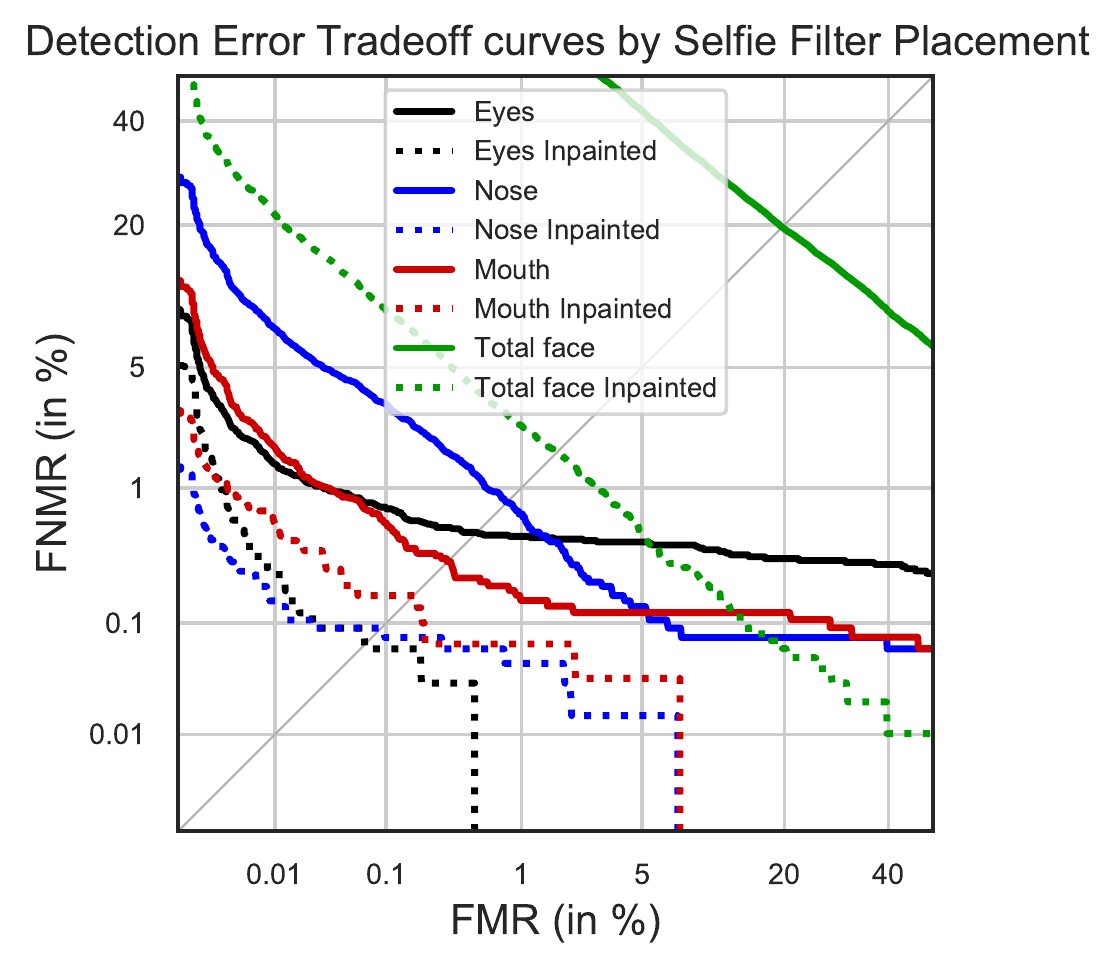}
  \caption{COTS}
  \label{_res222}
\end{subfigure}\quad %
\caption{DET curves for FRGC based selfie filter test dataset by Selfie Filter Placement.}
\label{_res22}
\vspace{-5pt}
\end{figure}

The application of the selfie filter removal when the entire face is occluded sees the highest FR performance enhancement but it can also improve performance in other scenarios as, for instance, illustrated in Fig.~\ref{_res22} and Tab.~\ref{FTE_result2}.

\begin{table}[h!]
\centering
\caption{EER comparison between the selfie filtered and its selfie filter removal counterpart by facial placement.}
\label{GAN_result2}
\begin{adjustbox}{max width=\columnwidth}
\begin{tabular}{@{\extracolsep{2pt}}llccc@{}} \toprule 
\multicolumn{1}{c}{}  &  & \multicolumn{1}{c}{\textbf{Selfie Filtered}} & \multicolumn{1}{c}{\textbf{\begin{tabular}[c]{@{}c@{}}Selfie Filter Removal\end{tabular}}} &  \\ 
\multirow{-2}{*}{\textbf{FR System}} & \multirow{-2}{*}{\textbf{Coverage}} &
\textbf{EER \%} &  \textbf{EER \%}  & \multirow{-2}{*}{\textbf{$\Delta$}}  \\ \midrule
ArcFace & Mouth & 0.820 & 3.100 & 2.280 \\ 
& Eyes  & 0.650 & 0.100 & \textbf{-0.550} \\ 
& Nose & 0.850 & 0.360 & \textbf{-0.490} \\ 
& Total face & 22.90 & 4.700 & \textbf{-18.20} \\ \midrule
COTS & Mouth & 0.500 & 0.250 & \textbf{-0.250} \\ 
& Eyes & 0.760 & 0.095 & \textbf{-0.665} \\ 
& Nose & 0.900 & 0.100 & \textbf{-0.800} \\
& Total face & 19.80 & 2.400 & \textbf{-17.40} \\
\bottomrule
\end{tabular}
\end{adjustbox}
\end{table}

\begin{table}[h!]
\centering
\caption{FTE comparison between the selfie filtered and its selfie filter removal counterpart by facial coverage.}
\label{FTE_result1}
\begin{adjustbox}{max width=\columnwidth}
\begin{tabular}{@{\extracolsep{2pt}}llccc@{}} \toprule 
\multicolumn{1}{c}{}  &  & \multicolumn{1}{c}{\textbf{Selfie Filtered}} & \multicolumn{1}{c}{\textbf{\begin{tabular}[c]{@{}c@{}}Selfie Filter Removal\end{tabular}}} &  \\ 
\multirow{-2}{*}{\textbf{FR System}} & \multirow{-2}{*}{\textbf{Coverage}} &
\textbf{FTE \%} &  \textbf{FTE \%}  & \multirow{-2}{*}{\textbf{$\Delta$}}  \\ \midrule
ArcFace & Low & 0.005 & 0.190 & 0.185 \\ 
& Medium & 0.015 & 0.278 & 0.263 \\ 
& High & 2.838 & 1.528 & \textbf{-1.310} \\ \midrule
COTS & Low & 0.319 & 0.502 & 0.183 \\ 
& Medium & 30.55 & 0.328 & \textbf{-30.22} \\ 
& High & 21.36 & 12.34 & \textbf{-9.02} \\
\bottomrule
\end{tabular}
\end{adjustbox}
\end{table}

In addition to reducing the EER, the selfie filter removal improves the FTE. Given that COTS has shown to be vulnerable to selfie filter covering the eye region, the selfie removal has reduced the corresponding FTE by 19.87\% points. Furthermore, the selfie filter removal has reduced the FTE on COTS for the total coverage selfie filters by 16.37\% points. The FTE metric on ArcFace is stable before and after the use of the selfie filter removal method. Tab.~\ref{FTE_result1} and Tab.~\ref{FTE_result2} highlight the benefits of the selfie filter removal method in terms of FTE across facial coverage and placement, where the FRGC test dataset is considered.

\begin{table}[h!]
\centering
\caption{FTE comparison between the selfie filtered and its selfie filter removal counterpart by facial placement.}
\label{FTE_result2}
\begin{adjustbox}{max width=\columnwidth}
\begin{tabular}{@{\extracolsep{2pt}}llccc@{}} \toprule 
\multicolumn{1}{c}{}  &  & \multicolumn{1}{c}{\textbf{Selfie Filtered}} & \multicolumn{1}{c}{\textbf{\begin{tabular}[c]{@{}c@{}}Selfie Filter Removal\end{tabular}}} &  \\ 
\multirow{-2}{*}{\textbf{FR System}} & \multirow{-2}{*}{\textbf{Coverage}} &
\textbf{FTE \%} &  \textbf{FTE \%}  & \multirow{-2}{*}{\textbf{$\Delta$}}  \\ \midrule
ArcFace & Mouth  & 0.244 & 1.917 & 1.673 \\ 
& Eyes & 0 & 0.208 & 0.208 \\ 
& Nose & 0.139 & 0 & \textbf{-0.139} \\ 
& Total face & 0.418 & 0.162 & \textbf{-0.256} \\ \midrule
COTS & Mouth & 0.662 & 1.439 & 0.777 \\ 
& Eyes & 20.66 & 0.786 & \textbf{-19.87} \\ 
& Nose & 0.069 & 0.494 & 0.425 \\
& Total face & 28.97 & 12.60 & \textbf{-16.37} \\
\bottomrule
\end{tabular}
\end{adjustbox}
\end{table}

\section{Conclusion}\label{sec:summary}
Fun filter are frequently used to modify selfies, \eg prior to sharing them on social media. Alterations and occlusions that are added to face images by applying such fun selfie filters represent a challenge for FR systems. The results obtained during this work have shown that fun selfie filters may negatively impact commercial and open-source FR modules. Across face detection, sample quality estimation, and FR, this is especially the case for facial images with high selfie filter facial coverage and for fun selfie filters that cover the mouth and nose. Furthermore, for the used COTS system, eye coverage has a high correlation with an increased FTE.

To tackle the above challenge, a selfie filter removal algorithm has been proposed. The proposed GAN-based method was shown to reduce the negative effects caused by the selfie filter when removing it prior to FR.

\section*{Acknowledgements}\label{sec:acknowledgement}
This research work has been partially funded by the German Federal Ministry of Education and Research and the Hessian Ministry of Higher Education, Research, Science and the Arts within their joint support of the National Research Center for Applied Cybersecurity ATHENE and the European Union’s Horizon 2020 research and innovation programme under the Marie Skłodowska-Curie grant agreement No. 860813 - TReSPAsS-ETN.

\bibliographystyle{IEEEtran.bst}
\bibliography{bibli}
\end{document}